\documentclass[11pt]{article}
\newcommand{\affilsup}[1]{\textsuperscript{#1}}
\usepackage{authblk}

\usepackage[final]{acl}
\usepackage{amsmath}
\usepackage{times}
\usepackage{latexsym}
\usepackage{multirow}
\usepackage{array}
\usepackage{booktabs}
\usepackage{tabularx}
\usepackage[capitalize,noabbrev]{cleveref}

\usepackage{algorithm}
\usepackage{algpseudocode}
\usepackage[utf8]{inputenc} %
\usepackage[T1]{fontenc}    %
\usepackage{hyperref}       %
\usepackage{url}            %
\usepackage{booktabs}       %
\usepackage{amsfonts}       %
\usepackage{nicefrac}       %
\usepackage{microtype}      %
\usepackage{xcolor}         %

\usepackage{subcaption}
\usepackage{hyperref}
\usepackage{wrapfig}
\usepackage{url}
\usepackage{paralist}
\usepackage{titletoc}

\usepackage[utf8]{inputenc}

\usepackage{microtype}

\usepackage{inconsolata}

\usepackage{graphicx}

\title{LadderEdit: Edit-Level Residual Compression for \\ Memory-Efficient Lifelong Editing of LLMs}

\author{Xiaobing Yu\affilsup{1,2} \quad Peijie Qiu\affilsup{2}\thanks{This work was done prior to joining Amazon.} \quad Jin Yang\affilsup{3} \quad Xuanzhao Dong\affilsup{4}\\
  Weiwei Ma\affilsup{2} \quad Zhaoqi An\affilsup{2} \quad Xiaoqi Zhao\affilsup{1}$^{\dag}$ \quad Xiaofeng Liu\affilsup{1}$^{\dag}$\\
  $^1$\large{Yale University}; \quad $^2$\large{Washington University in St. Louis}; \\
  \quad $^3$\large{Icahn School of Medicine at Mount Sinai}; \quad $^4$\large{Arizona State University}\\
  \texttt{\{xiaobing.yu, xiaoqi.zhao, xiaofeng.liu\}@yale.edu} \\
$^{\dag}$ co-corresponding authors\\
\textbf{Code}: \url{https://github.com/VisualReasoner/LadderEdit}
}

\begin{document}
\maketitle
\begin{abstract}
Lifelong editing of LLMs requires storing thousands of edits after acquisition. 
A widely used family of approaches attaches one LoRA adapter per edit, which preserves behavior but grows linearly in storage. To address this challenge, we propose LadderEdit, a method that compresses each LoRA adapter after it is acquired. Each edit is first stored at low rank as a cheap sketch. We then check whether this sketch still satisfies the rewrite, generalization, and locality contract on probe prompts. Edits that pass keep the sketch; those that fail are promoted to a higher rank along a ladder until the contract is met. Because every edit retains some representation, coverage is maintained, and only hard edits consume more rank. Across ZsRE, CounterFact, and WikiBigEdit benchmarks on LLaMA-3-8B, Mistral-7B, and Qwen2.5-7B, LadderEdit tracks exact LoRA storage at $5.2\times$ less memory and remains effective at $50{,}000$ sequential edits.
\end{abstract}

\section{Introduction}
\label{sec:intro}

Lifelong model editing asks a deployed large language model to absorb a stream of factual corrections and knowledge updates without full retraining. This setting is increasingly important: world knowledge changes, model errors are discovered after deployment, and domain-specific assistants require controlled updates over time. A useful lifelong editor must therefore satisfy three requirements simultaneously. It must remain plastic enough to incorporate new edits, stable enough to preserve previous edits and unrelated knowledge, and efficient enough to serve after thousands of updates.

Existing approaches expose a fundamental storage-plasticity tension. Shared-parameter and constant-memory editors bound the deployed memory footprint, but every new edit competes with previous edits for limited capacity~\cite{grace,wang2026memoir}. As the stream grows, such methods can preserve locality while losing edit plasticity: the model avoids damaging unrelated knowledge, but new edits may fail to generalize beyond the rewrite prompt. Retrieval- and routing-based editors reduce direct parameter interference but introduce a separate edit-selection problem and can fail when a query is semantically close to, but not exactly matched with, the stored edit~\cite{elder,memoe,adalora,lemoe}. Direct parameter-editing methods avoid explicit per-edit storage, yet repeated updates can interfere destructively over long edit streams~\cite{memit,rome,decao2021editing,dai2022knowledge,fang2025alphaedit,easyedit,mend}.

\begin{figure}[t]
    \centering
    \includegraphics[width=1.0\columnwidth]{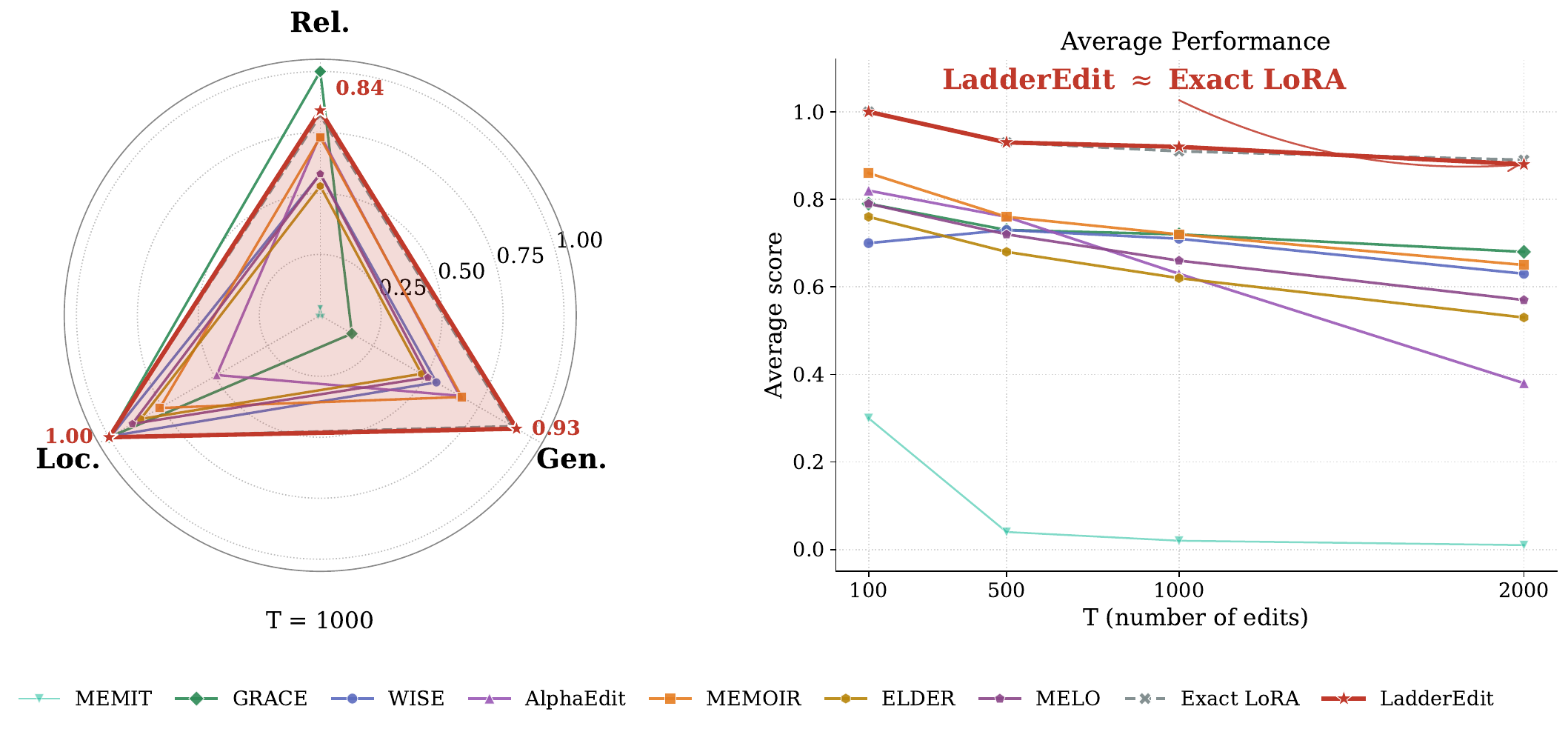}
    \caption{Performance comparison of LadderEdit, exact LoRA, and baseline methods across sequential editing tasks. LadderEdit usually tracks exact LoRA closely while reducing persistent memory, with small losses in some compressed settings.}
    \label{fig:editing-comparison}
\end{figure}

An alternative is to store edit-local update memories. Each edit is written into an isolated update representation, so later edits do not overwrite earlier ones~\cite{hu2022lora,melo}. This design preserves edit-specific plasticity and provides a clean full-storage reference: if the correct edit memory is selected, the system can recover the behavior of the original update. The drawback is persistent storage. If every edit is retained at full resolution, memory grows linearly with the number of edits. Thus, the bottleneck in long-horizon editing is not only how to acquire an edit but also how much of each acquired edit must remain stored.

To address this dilemma, we reframe lifelong edit storage as an edit-level resolution allocation problem. The key is not simply which edits should be kept or dropped but how much representational resolution each edit needs to preserve its behavioral contract. This distinction matters because keep/drop caching couples memory reduction to edit absence: once an edit is dropped, it has no representation and cannot support rewrite, generalization, or locality. In contrast, a resolution-allocation policy can preserve coverage for every edit through a cheap representation, while spending additional memory only on edits that require higher fidelity.

We propose \textsc{LadderEdit}, a storage-layer method based on this coverage-before-fidelity principle. Each incoming edit is first acquired as a full-resolution update memory; in our implementation, this update memory is an exact LoRA update. LadderEdit is a two-stage controller. A \emph{spectral predictor} scores each candidate rank using the singular-value tail of the update together with the edit's behavioral slack, and proposes the cheapest rank $\hat{r}_i$ whose score falls below a calibrated threshold; this proposal requires only one SVD per edit and no behavioral evaluation. A \emph{behavioral audit} then verifies the proposed sketch against rewrite, generalization, and locality probes. If the audit passes, the edit is stored at $\hat{r}_i$; if it fails, the audit promotes the edit along the spectrally-ordered ladder until a passing rung is found. This design avoids exhaustive auditing across all rungs and gives every edit coverage through its predicted rung while spending additional resolution only when audit demands it.

We validate on ZsRE~\cite{levy2017zero}, CounterFact (CF)~\cite{rome,memit}, and WikiBigEdit~\cite{thede2025wikibigedit} using LLaMA-3-8B~\cite{LLaMA3}, Mistral-7B~\cite{jiang2023mistral}, and Qwen2.5-7B~\cite{qwen2025,bai2025qwen3,yang2025qwen3}. Figure~\ref{fig:editing-comparison} shows LadderEdit tracks the full-resolution reference at a fraction of the memory, and budget-matched frontiers confirm that residual-ladder allocation substantially outperforms exact-cache: preserving coverage before fidelity is the better storage primitive for scalable lifelong editing.

Our contributions are as follows: 
\emph{i)} We formulate lifelong editing storage as a resolution-allocation problem, separating edit acquisition, selection, and persistent representation. \emph{ii)} We introduce LadderEdit, a residual-ladder controller that covers every edit with a low-rank sketch and restores detail only when behavioral audits demand it. \emph{iii)} We identify \emph{coverage-before-fidelity} as the principle for scalable edit memory, nearly doubling utility over exact-cache at matched budget. \emph{iv} We validate LadderEdit on ZsRE, CF, and WikiBigEdit, tracking full-resolution memory at $5.2\times$ less storage and remaining effective at 50{,}000 edits.


\begin{figure*}[t]
    \centering
    \includegraphics[width=0.95\textwidth]{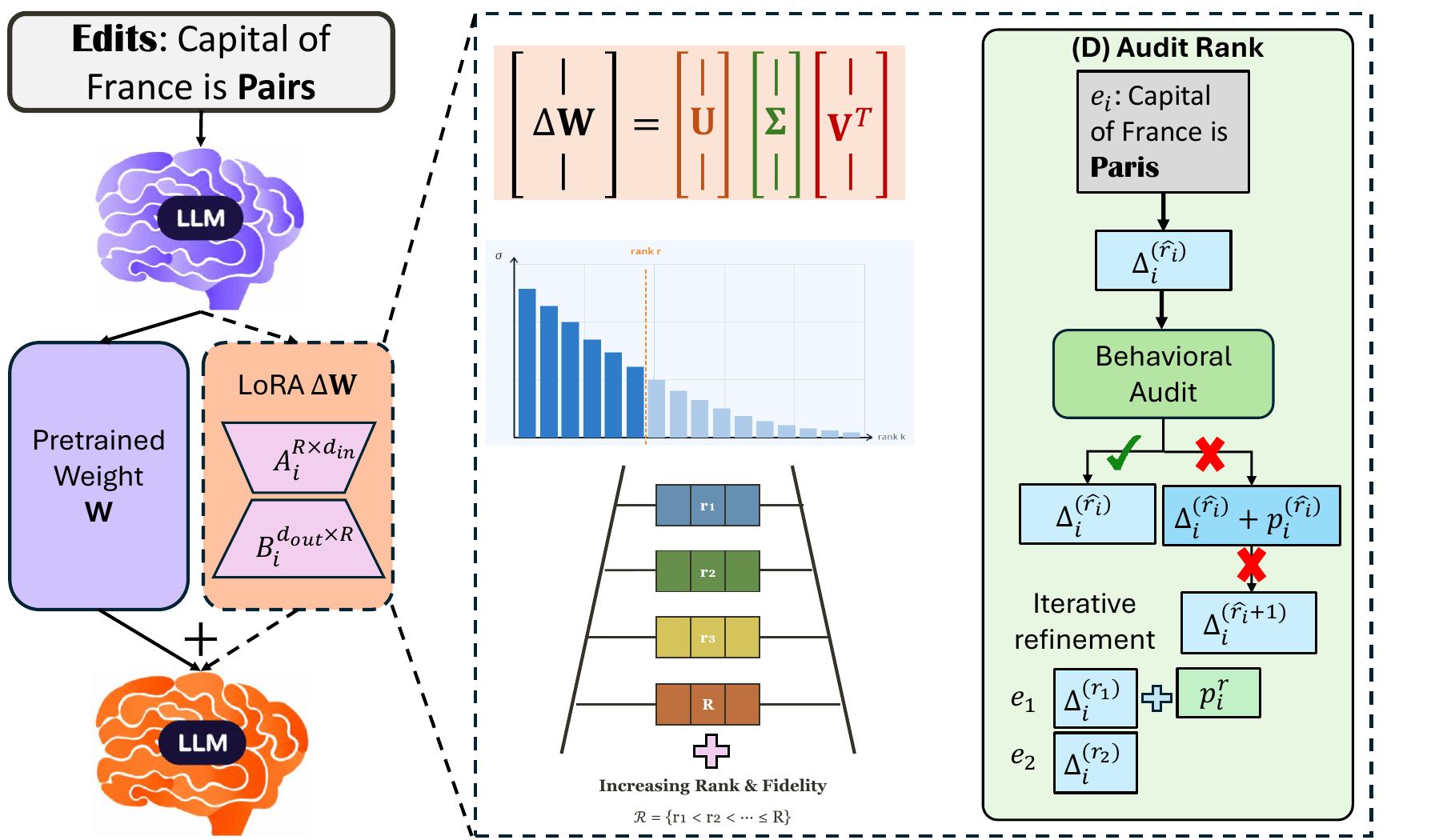}
    \caption{LadderEdit insertion pipeline. We first fit an exact LoRA update for the incoming edit, then a candidate rank $\hat{r}_i$ from the update spectrum and exact-edit margin (one SVD, no behavioral cost). We further decompose the edit into ladder rungs: low-rank sketches at increasing rank. And finally we audit the predicted sketch against rewrite, generalization, and locality. If the audit passes, store at $\hat{r}_i$; otherwise promote along the spectral ladder until a passing rung is found.}
    \label{fig:method-overview}
\end{figure*}

\section{Related Work}
\label{sec:related}

\paragraph{Model editing and lifelong editing.}
Model editing methods update a model's behavior without full retraining, either by directly modifying parameters~\cite{sinitsin2020editable,decao2021editing,rome,memit}, learning editor networks~\cite{mend}, or routing edited cases through external/contextual memories~\cite{serac,ike}. These methods establish the basic reliability, generalization, and locality criteria, but most are not designed for long edit streams where thousands of updates must coexist. Lifelong editing methods address this sequential setting through key-value memories, side memories, null-space constraints, or adapter/MoE-based mechanisms~\cite{grace,wise,fang2025alphaedit,wang2026memoir,melo,elder,memoe,lemoe}. However, they still face a storage-plasticity tension: shared or constant-memory systems can lose plasticity as capacity saturates, while edit-local memories preserve behavior but grow with the number of edits. LadderEdit targets this storage layer directly.

\paragraph{Low-rank adaptation and edit-memory storage.}
LoRA and its variants show that low-rank updates provide an efficient representation for adapting large models~\cite{hu2022lora,dettmers2023qlora,dora,reloRA,yu2025fm}. Adaptive-rank methods such as AdaLoRA~\cite{adalora} allocate rank across layers or modules during training, while serving systems such as S-LoRA~\cite{slora} study how to deploy many adapters efficiently. LadderEdit addresses a different problem: after an edit has already been acquired, how much of its update memory must be persistently stored? The unit of allocation is an edit rather than a layer, and the criterion is behavioral contract satisfaction rather than gradient-derived importance. Thus LadderEdit is not ordinary LoRA compression or training-time rank allocation; it is an edit-level resolution-allocation method that keeps every edit covered by a sketch and promotes to higher rank only when needed.

\paragraph{Evaluation under compression.}
Recent surveys and evaluations emphasize that edited models should be judged not only by rewrite success but also by generalization, locality, long-form behavior, and degradation over many edits~\cite{yao2023editing,zhang2024comprehensive,wang2024survey,longformeval,reallyedit,lifelongllmsurvey,contllmsurvey}. LadderEdit follows this view at the storage level: spectral compressibility proposes candidate rungs, but the final decision is made by behavioral auditing over rewrite, generalization, and locality. This is the central difference from exact-cache allocation, which reduces memory by dropping edits rather than by allocating resolution.

LadderEdit's formulation differs from ordinary LoRA compression and adaptive rank training. LadderEdit does not learn a better adapter at acquisition time and does not solve retrieval or task identification; it operates after the edit has been acquired, inheriting the selection mechanism of the full-resolution memory bank, and asks a storage-layer question:
once an edit is selected, can the system return a cheaper representation without violating the measured edit contract? The advantage comes from changing the memory primitive. Exact-cache
allocation saves memory by storing a subset of edits exactly and dropping the rest, reducing coverage. LadderEdit first
gives every edit coverage through a sketch, then allocates fidelity through additional rank only where the audit shows it is
needed. The two policies fail differently under compression: exact-cache fails by absence, whereas LadderEdit degrades
by resolution. A focused comparison that distinguishes LadderEdit from
training-time rank allocation~\cite{adalora}, multi-adapter serving~\cite{slora}, and adapter merging is provided in
Appendix~\ref{app:why-not-standard}.

\section{Method}
\label{sec:method}

\subsection{Problem Setting}
\label{sec:problem-setting}

We consider lifelong editing for a pretrained LLM $f_\theta$, where edits arrive as a sequence $\{e_i\}_{i=1}^N$. After each insertion, the system must answer the edited prompt correctly, generalize the edit to related prompts, and preserve unrelated knowledge. We refer to these requirements as reliability, generalization, and locality. The scaling challenge is not only to satisfy this behavioral contract for one edit, but to keep satisfying it as $N$ grows.

We instantiate each edit with a LoRA-style update. Let $\Delta_i^E$ denote
the exact update produced by the edit writer for edit $e_i$. Two extremes
bracket the space of persistent representations: a full-resolution
edit-local memory bank stores every $\Delta_i^E$ exactly, scaling linearly
as $NP_E$ where $P_E$ is the cost of one exact edit; a shared or
constant-memory editor folds all edits into a fixed footprint, but
plasticity degrades as later edits compete with earlier ones for finite
capacity. Neither extreme matches the empirical structure of an edit
stream: across $N = 2{,}000$ edits on both ZsRE and CounterFact, roughly
$80\%$ of edits are behaviorally sketch-sufficient at rank~1, a
$\sim\!19\%$ hard tail requires rank~2 or higher, and a
smaller $\sim\!5\%$ population is \emph{compression-regularized}, the
rank-1 sketch satisfies the behavioral contract that the exact update
fails (Appendix~\ref{app:heterogeneity}). LadderEdit occupies the
intermediate regime that this distribution implies: every edit retains
its own representation, but that representation may be a compressed rung
rather than the full exact adapter. Table~\ref{tab:heterogeneity} reports the population structure of
the edit stream that motivates this design.

The system rests on three commitments. It is \emph{exact first}: each edit is acquired at full resolution by the standard writer, so the acquired update is preserved without loss. It is \emph{retrieval agnostic}: it inherits whatever selection mechanism the deployment uses, with the storage gain preserved under a learned sentence encoder retriever (Appendix~\ref{app:appendix_retrieval}). It is \emph{contract grounded}: at insertion, a behavioral audit certifies that the returned representation satisfies the same rewrite, generalization, and locality test as $\Delta_i^E$. At inference, retrieval selects the edit and LadderEdit returns the cheapest audited rung; the acquisition cost of $\Delta_i^E$ is paid once, and the persistent cost is whatever the audit permits. LadderEdit decides which directions of $\Delta_i^E$ to retain and composes with quantization, which acts on bit width (Appendix~\ref{app:quantization}).

\subsection{Residual-Ladder Allocation}

LadderEdit is a residual-ladder allocation mechanism for scaling edit-local update memories in lifelong editing. It keeps the isolation and plasticity of per-edit adapters, but stores most edits at lower resolution. The method has three steps: acquire an exact update, compile that update into candidate rungs, and audit the cheapest safe rung. Figure~\ref{fig:method-overview} summarizes the full insertion-time pipeline, and Algorithm \ref{alg:ladderedit} gives the controller.

For an incoming edit $e_i$, let $\Delta_i^E$ denote the exact LoRA-style update acquired by the edit writer. LadderEdit constructs a rank-$r$ sketch:
\begin{equation}
\label{eq:rank-sketch}
\Delta_i^{(r)} = \Pi_r(\Delta_i^E),
\end{equation}
We use the rank menu $\mathcal{R} = \{1, 2, \ldots, R_{\max}\}$, matching the same acquisition rank as $R_{\max}$ from the exact LoRA update.

where $\Pi_r$ truncates the singular directions of the effective LoRA update. The omitted detail is the residual:
\begin{equation}
\label{eq:residual}
\rho_i^{(r)} = \Delta_i^E - \Delta_i^{(r)}.
\end{equation}
The deployed edit is one rung of the ladder:
\begin{equation}
\label{eq:deployed-edit}
\Delta_i = \Delta_i^{(r_i)}.
\end{equation}
When $r_i=R_{\max}$, the stored update equals the exact LoRA edit. The key difference from exact-cache allocation is that every edit keeps a representation, and memory is spent on resolution rather than on a binary keep/drop decision.

Rank selection starts from a cheap spectral proposal. Let $T_i(r)$ be the normalized spectral tail after rank $r$. We define the exact-edit margin as the minimum behavioral slack of the full-resolution edit:
\begin{equation}
\label{eq:score}
\mu_i^E=\min\{R_i^E-\tau_R,\;G_i^E-\tau_G,\;L_i^E-\tau_L\}.
\end{equation}
Positive margin means the acquired edit satisfies all three measured constraints, and larger margin indicates more tolerance for compression.

LadderEdit scores:
\begin{equation}
\label{eq:compressibility-score}
S_i(r) =
\frac{T_i(r)}{[\mu_i^E]_+ + \epsilon},
\end{equation}
where $\epsilon>0$ prevents unstable division. Lower scores indicate that little spectral mass remains outside the sketch relative to the exact edit's behavioral slack. The cheapest rank whose score falls below a calibrated threshold $\tau$ is the predicted rank $\hat{r}_i$:
\begin{equation}
\hat{r}_i = \min\{ r \in \mathcal{R} : S_i(r) \leq \tau \}.
\end{equation}
The spectral score is a proposal, not a safety proof; it selects \emph{which single rung} the audit will inspect first, so the audit sees one candidate per edit instead of $|\mathcal{R}|$ candidates.

Behavioral auditing remains the final gate. For a candidate stored representation $m$, let $R_i^m$, $G_i^m$, and $L_i^m$ denote rewrite, generalization, and locality scores. We define the soft contract gap:
\begin{equation}
\label{eq:contract-gap}
\begin{aligned}
H_i^m ={}&
w_R[\tau_R-R_i^m]_+
+w_G[\tau_G-G_i^m]_+ \\
&+w_L[\tau_L-L_i^m]_+ .
\end{aligned}
\end{equation}
The audit score in Eq.~\ref{eq:contract-gap} uses weights $w_R=1.0$, $w_G=0.8$, and $w_L=1.2$. Locality receives the highest weight because locality failures are catastrophic and irreversible, while reliability and generalization failures trigger rank promotion and recover within the same edit. We swept $w_L$ over $\{0.8, 1.0, 1.2, 1.5, 2.0\}$ on the validation split and Avg.\ stayed within 0.01 of the default for $w_L\in[1.0,1.5]$ and dropped by 0.02 and 0.03 at the endpoints 0.8 and 2.0 respectively. The default $w_L=1.2$ sits at the midpoint of the stable region. The weights $w_R$ and $w_G$ follow the standard LoRA loss-weighting ratio. We demonstrate more details in Appendix~\ref{app:audit-weights}.

\paragraph{Downgrade policy.} Algorithm~\ref{alg:ladderedit} includes budget-based downgrading. When the audit succeeds at rank $r$, the controller checks whether the rank-$(r{-}1)$ approximation $U_{:,1:r-1}\Sigma_{1:r-1,1:r-1}V_{1:r-1,:}^T$ captures more than 92\% of the Frobenius energy of the full LoRA update. If so, the controller commits the rank-$(r{-}1)$ sketch and re-runs the audit on the downgraded sketch, and if the downgrade also passes the audit, it is kept. Downgrade attempts are bounded at one per edit to keep insertion cost bounded. Downgrades occur for 3.1\% of edits at $T=1{,}000$ on LLaMA-3-8B / ZsRE and save approximately 0.6~GB of persistent memory at $T=10{,}000$. The 0.92 Frobenius threshold was selected on the validation split, and varying it from 0.88 to 0.95 changes Avg.\ by less than 0.005.

If $H_i^m=0$, the candidate satisfies the edit contract. If the audit at $\hat{r}_i$ fails ($H_i^{(\hat{r}_i)} > 0$), LadderEdit re-audits at the next rung along the spectrally-ordered ladder $\hat{r}_i + 1, \hat{r}_i + 2, \ldots$ until the contract is met or $R_{\max}$ is reached. In practice the predicted rung passes on the first audit for the majority of edits (Appendix~\ref{app:rung-allocation}), so the expected number of audits per edit is close to one.

Under a memory budget, higher-rank promotion is allocated where it removes the most contract violation per parameter. Let $P_E$ be the exact edit cost and $P_r$ the rank-$r$ sketch cost. The promotion value density is:
\begin{equation}
\label{eq:rvd}
\mathrm{RVD}_i(r) =
\frac{H_i^{(r)} - H_i^E}{P_E-P_r}.
\end{equation}
If all $N$ edits use rank $r$ and $K$ edits are promoted to the exact rung, persistent memory is:
\begin{equation}
\label{eq:memory}
M(r,K)=NP_r+K(P_E-P_r).
\end{equation}
For analytical clarity, Eq.~\ref{eq:rvd}--\ref{eq:memory} use a two-state limit in which promoted edits jump directly to $R_{\max}$. The deployed controller supports intermediate promotion along the full ladder $\mathcal{R}$.
The first term gives every edit a cheap plastic representation. The second term spends exact memory only on the difficult subset. This yields a practical route for one-LoRA-per-edit scaling whenever $K \ll N$.

\begin{algorithm}[t]
\caption{\textsc{LadderEdit}: spectral-prediction with audited promotion}
\label{alg:ladderedit}
\footnotesize
\begin{algorithmic}[1]
\Require Edit stream $\{e_i\}_{i=1}^N$, frozen model $f_\theta$, rank menu $\mathcal{R}=\{1,\ldots,R_{\max}\}$, threshold $\tau$, memory budget $B$
\State Initialize stored memory $\mathcal{M} \gets \emptyset$
\For{each incoming edit $e_i$}
    \State Acquire exact LoRA update $\Delta_i^E$
    \State Compute SVD of $\Delta_i^E$ and form spectral tails $T_i(r)$ for all $r \in \mathcal{R}$
    \State Estimate exact-edit margin $\mu_i^E$ on rewrite, generalization, and locality probes
    \State \textbf{Predict:} $\hat{r}_i \gets \min\{ r \in \mathcal{R} : S_i(r) \leq \tau \}$ \Comment{single proposal, no behavioral cost}
    \State $r_i \gets \hat{r}_i$
    \State \textbf{Audit:} build sketch $\Delta_i^{(r_i)}$, compute contract gap $H_i^{(r_i)}$
    \While{$H_i^{(r_i)} > 0$ \textbf{and} $r_i < R_{\max}$} \Comment{audited promotion}
        \State $r_i \gets r_i + 1$
        \State Rebuild sketch $\Delta_i^{(r_i)}$, recompute $H_i^{(r_i)}$
    \EndWhile
    \State Store $\Delta_i^{(r_i)}$ in $\mathcal{M}$
    \If{stored memory exceeds $B$}
        \State Downgrade edits with lowest value density
    \EndIf
\EndFor
\State \Return stored edit memory $\mathcal{M}$
\end{algorithmic}
\end{algorithm}

The rank-analysis package also clarifies how to interpret storage rungs. An edit is \emph{sketch-sufficient} if both sketch and exact pass, \emph{detail-needed} if the sketch fails but exact passes, \emph{compression-regularized} if the sketch passes but exact fails, and \emph{exact-limited} if both fail. More details in Appendix \ref{app:appendix-examples}.

\section{Experiments}
\label{sec:experimental}
\subsection{Experimental Setup}
We evaluate LadderEdit as a storage-layer method for edit-local memories. The central question is whether lifelong edit storage should be allocated by \emph{coverage} and \emph{fidelity}, rather than by exact keep/drop caching. We therefore compare exact LoRA as the full-resolution reference, fixed-rank compression as a nonadaptive baseline, exact-cache as a keep/drop policy, and LadderEdit as an edit-level resolution-allocation policy across standard sequential editing benchmarks.

\begin{table*}[t]
  \caption{Q\&A task results on the ZsRE dataset. $T$ denotes the number of edits. Mean $\pm$ standard deviation over three random seeds. Avg.\ is the arithmetic mean of Rel., Gen., and Loc. The best average performance per column is marked in \textbf{bold}.}
  \vspace{3pt}
  \centering
  \resizebox{1.0\linewidth}{!}{%
  \begin{tabular}{l|cccc|cccc|cccc|cccc}
    \toprule
    \multirow{2}{*}{\textbf{Method}}
      & \multicolumn{4}{c|}{$T=100$}
      & \multicolumn{4}{c|}{$T=500$}
      & \multicolumn{4}{c|}{$T=1{,}000$}
      & \multicolumn{4}{c}{$T=2{,}000$} \\
    \cmidrule(lr){2-5}\cmidrule(lr){6-9}\cmidrule(lr){10-13}\cmidrule(lr){14-17}
      & Rel. & Gen. & Loc. & Avg.
      & Rel. & Gen. & Loc. & Avg.
      & Rel. & Gen. & Loc. & Avg.
      & Rel. & Gen. & Loc. & Avg. \\
    \midrule
    & \multicolumn{16}{c}{\texttt{\textbf{LLaMA-3-8B}}} \\
    \midrule
    FT & 0.13\,\tiny$\pm$.004&0.11\,\tiny$\pm$.005&0.02\,\tiny$\pm$.001&0.09\,\tiny$\pm$.002 & 0.14\,\tiny$\pm$.005&0.12\,\tiny$\pm$.006&0.02\,\tiny$\pm$.001&0.09\,\tiny$\pm$.003 & 0.13\,\tiny$\pm$.005&0.12\,\tiny$\pm$.007&0.01\,\tiny$\pm$.001&0.09\,\tiny$\pm$.003 & 0.09\,\tiny$\pm$.006&0.08\,\tiny$\pm$.007&0.01\,\tiny$\pm$.001&0.06\,\tiny$\pm$.003 \\
    ROME~\cite{rome} & 0.08\,\tiny$\pm$.003&0.08\,\tiny$\pm$.003&0.02\,\tiny$\pm$.001&0.06\,\tiny$\pm$.001 & 0.04\,\tiny$\pm$.002&0.04\,\tiny$\pm$.002&0.02\,\tiny$\pm$.001&0.03\,\tiny$\pm$.001 & 0.03\,\tiny$\pm$.002&0.03\,\tiny$\pm$.002&0.02\,\tiny$\pm$.001&0.03\,\tiny$\pm$.001 & 0.01\,\tiny$\pm$.001&0.01\,\tiny$\pm$.001&0.01\,\tiny$\pm$.001&0.01\,\tiny$\pm$.001 \\
    MEMIT~\cite{memit} & 0.03\,\tiny$\pm$.002&0.03\,\tiny$\pm$.002&0.01\,\tiny$\pm$.001&0.02\,\tiny$\pm$.001 & 0.01\,\tiny$\pm$.001&0.01\,\tiny$\pm$.001&0.00\,\tiny$\pm$.000&0.01\,\tiny$\pm$.001 & 0.00\,\tiny$\pm$.000&0.00\,\tiny$\pm$.000&0.00\,\tiny$\pm$.000&0.00\,\tiny$\pm$.001 & 0.00\,\tiny$\pm$.000&0.00\,\tiny$\pm$.000&0.00\,\tiny$\pm$.000&0.00\,\tiny$\pm$.001 \\
    GRACE~\cite{grace} & 1.00\,\tiny$\pm$.001&0.39\,\tiny$\pm$.009&1.00\,\tiny$\pm$.001&0.80\,\tiny$\pm$.003 & 1.00\,\tiny$\pm$.001&0.38\,\tiny$\pm$.011&1.00\,\tiny$\pm$.001&0.79\,\tiny$\pm$.004 & 1.00\,\tiny$\pm$.001&0.37\,\tiny$\pm$.012&1.00\,\tiny$\pm$.001&0.79\,\tiny$\pm$.004 & 1.00\,\tiny$\pm$.001&0.34\,\tiny$\pm$.013&1.00\,\tiny$\pm$.001&0.78\,\tiny$\pm$.004 \\
    WISE~\cite{wise} & 0.62\,\tiny$\pm$.014&0.60\,\tiny$\pm$.015&1.00\,\tiny$\pm$.001&0.74\,\tiny$\pm$.007 & 0.69\,\tiny$\pm$.017&0.66\,\tiny$\pm$.018&1.00\,\tiny$\pm$.001&0.78\,\tiny$\pm$.008 & 0.66\,\tiny$\pm$.019&0.64\,\tiny$\pm$.020&1.00\,\tiny$\pm$.001&0.77\,\tiny$\pm$.009 & 0.55\,\tiny$\pm$.021&0.52\,\tiny$\pm$.022&1.00\,\tiny$\pm$.001&0.69\,\tiny$\pm$.010 \\
    AlphaEdit~\cite{fang2025alphaedit} & 0.91\,\tiny$\pm$.012&0.79\,\tiny$\pm$.013&0.94\,\tiny$\pm$.008&0.88\,\tiny$\pm$.006 & 0.89\,\tiny$\pm$.014&0.78\,\tiny$\pm$.016&0.81\,\tiny$\pm$.010&0.83\,\tiny$\pm$.008 & 0.84\,\tiny$\pm$.016&0.77\,\tiny$\pm$.018&0.56\,\tiny$\pm$.011&0.72\,\tiny$\pm$.009 & 0.62\,\tiny$\pm$.018&0.58\,\tiny$\pm$.019&0.21\,\tiny$\pm$.012&0.47\,\tiny$\pm$.010 \\
    MEMOIR~\cite{wang2026memoir} & 0.87\,\tiny$\pm$.008&0.84\,\tiny$\pm$.010&0.95\,\tiny$\pm$.006&0.89\,\tiny$\pm$.005 & 0.82\,\tiny$\pm$.010&0.75\,\tiny$\pm$.012&0.82\,\tiny$\pm$.007&0.80\,\tiny$\pm$.006 & 0.77\,\tiny$\pm$.011&0.71\,\tiny$\pm$.014&0.80\,\tiny$\pm$.008&0.76\,\tiny$\pm$.007 & 0.71\,\tiny$\pm$.012&0.62\,\tiny$\pm$.015&0.77\,\tiny$\pm$.009&0.70\,\tiny$\pm$.007 \\
    ELDER~\cite{elder} & 0.78\,\tiny$\pm$.014&0.72\,\tiny$\pm$.016&0.93\,\tiny$\pm$.009&0.81\,\tiny$\pm$.008 & 0.69\,\tiny$\pm$.017&0.63\,\tiny$\pm$.019&0.91\,\tiny$\pm$.011&0.74\,\tiny$\pm$.009 & 0.61\,\tiny$\pm$.019&0.55\,\tiny$\pm$.022&0.89\,\tiny$\pm$.012&0.68\,\tiny$\pm$.010 & 0.49\,\tiny$\pm$.021&0.44\,\tiny$\pm$.024&0.86\,\tiny$\pm$.013&0.60\,\tiny$\pm$.011 \\
    MELO~\cite{melo} & 0.81\,\tiny$\pm$.012&0.76\,\tiny$\pm$.014&0.96\,\tiny$\pm$.008&0.84\,\tiny$\pm$.007 & 0.73\,\tiny$\pm$.014&0.67\,\tiny$\pm$.017&0.94\,\tiny$\pm$.010&0.78\,\tiny$\pm$.008 & 0.66\,\tiny$\pm$.016&0.59\,\tiny$\pm$.019&0.92\,\tiny$\pm$.011&0.72\,\tiny$\pm$.009 & 0.55\,\tiny$\pm$.018&0.49\,\tiny$\pm$.021&0.89\,\tiny$\pm$.012&0.64\,\tiny$\pm$.010 \\
    AdaLoRA~\cite{adalora} & 0.65\,\tiny$\pm$.020&0.55\,\tiny$\pm$.022&0.20\,\tiny$\pm$.010&0.47\,\tiny$\pm$.010 & 0.01\,\tiny$\pm$.001&0.01\,\tiny$\pm$.001&0.00\,\tiny$\pm$.000&0.01\,\tiny$\pm$.001 & 0.00\,\tiny$\pm$.000&0.00\,\tiny$\pm$.000&0.00\,\tiny$\pm$.000&0.00\,\tiny$\pm$.001 & 0.00\,\tiny$\pm$.000&0.00\,\tiny$\pm$.000&0.00\,\tiny$\pm$.000&0.00\,\tiny$\pm$.001 \\
    \midrule
    LoRA (Exact)~\cite{hu2022lora} & 1.00\,\tiny$\pm$.001&0.99\,\tiny$\pm$.002&1.00\,\tiny$\pm$.001&\textbf{1.00\,\tiny$\pm$.001} & 0.85\,\tiny$\pm$.006&0.94\,\tiny$\pm$.007&1.00\,\tiny$\pm$.001&0.93\,\tiny$\pm$.003 & 0.95\,\tiny$\pm$.007&0.91\,\tiny$\pm$.008&1.00\,\tiny$\pm$.001&\textbf{0.95\,\tiny$\pm$.004} & 0.94\,\tiny$\pm$.007&0.89\,\tiny$\pm$.009&1.00\,\tiny$\pm$.001&\textbf{0.94\,\tiny$\pm$.004} \\
    \textbf{LadderEdit (Ours)} & 1.00\,\tiny$\pm$.001&0.99\,\tiny$\pm$.002&1.00\,\tiny$\pm$.001&\textbf{1.00\,\tiny$\pm$.001} & 0.97\,\tiny$\pm$.005&0.94\,\tiny$\pm$.006&1.00\,\tiny$\pm$.001&\textbf{0.97\,\tiny$\pm$.003} & 0.95\,\tiny$\pm$.005&0.91\,\tiny$\pm$.007&1.00\,\tiny$\pm$.001&\textbf{0.95\,\tiny$\pm$.003} & 0.94\,\tiny$\pm$.006&0.89\,\tiny$\pm$.007&1.00\,\tiny$\pm$.001&\textbf{0.94\,\tiny$\pm$.003} \\
    \midrule
    & \multicolumn{16}{c}{\texttt{\textbf{Mistral-7B}}} \\
    \midrule
    FT & 0.11\,\tiny$\pm$.005&0.10\,\tiny$\pm$.006&0.02\,\tiny$\pm$.001&0.08\,\tiny$\pm$.003 & 0.15\,\tiny$\pm$.006&0.13\,\tiny$\pm$.007&0.02\,\tiny$\pm$.001&0.10\,\tiny$\pm$.003 & 0.16\,\tiny$\pm$.006&0.13\,\tiny$\pm$.008&0.01\,\tiny$\pm$.001&0.10\,\tiny$\pm$.003 & 0.08\,\tiny$\pm$.007&0.07\,\tiny$\pm$.009&0.01\,\tiny$\pm$.001&0.05\,\tiny$\pm$.004 \\
    ROME~\cite{rome} & 0.05\,\tiny$\pm$.002&0.05\,\tiny$\pm$.002&0.02\,\tiny$\pm$.001&0.04\,\tiny$\pm$.001 & 0.04\,\tiny$\pm$.002&0.04\,\tiny$\pm$.002&0.02\,\tiny$\pm$.001&0.03\,\tiny$\pm$.001 & 0.04\,\tiny$\pm$.002&0.04\,\tiny$\pm$.002&0.02\,\tiny$\pm$.001&0.03\,\tiny$\pm$.001 & 0.01\,\tiny$\pm$.001&0.01\,\tiny$\pm$.001&0.01\,\tiny$\pm$.001&0.01\,\tiny$\pm$.001 \\
    MEMIT~\cite{memit} & 0.00\,\tiny$\pm$.000&0.00\,\tiny$\pm$.000&0.01\,\tiny$\pm$.001&0.00\,\tiny$\pm$.001 & 0.02\,\tiny$\pm$.001&0.02\,\tiny$\pm$.001&0.01\,\tiny$\pm$.001&0.02\,\tiny$\pm$.001 & 0.04\,\tiny$\pm$.002&0.04\,\tiny$\pm$.002&0.02\,\tiny$\pm$.001&0.03\,\tiny$\pm$.001 & 0.01\,\tiny$\pm$.001&0.01\,\tiny$\pm$.001&0.01\,\tiny$\pm$.001&0.01\,\tiny$\pm$.001 \\
    GRACE~\cite{grace} & 1.00\,\tiny$\pm$.001&0.15\,\tiny$\pm$.010&1.00\,\tiny$\pm$.001&0.72\,\tiny$\pm$.003 & 1.00\,\tiny$\pm$.001&0.09\,\tiny$\pm$.012&1.00\,\tiny$\pm$.001&0.70\,\tiny$\pm$.004 & 1.00\,\tiny$\pm$.001&0.02\,\tiny$\pm$.001&1.00\,\tiny$\pm$.001&0.67\,\tiny$\pm$.001 & 1.00\,\tiny$\pm$.001&0.01\,\tiny$\pm$.001&1.00\,\tiny$\pm$.001&0.67\,\tiny$\pm$.001 \\
    WISE~\cite{wise} & 0.87\,\tiny$\pm$.016&0.80\,\tiny$\pm$.017&1.00\,\tiny$\pm$.001&0.89\,\tiny$\pm$.008 & 0.80\,\tiny$\pm$.019&0.74\,\tiny$\pm$.021&1.00\,\tiny$\pm$.001&0.85\,\tiny$\pm$.009 & 0.70\,\tiny$\pm$.022&0.67\,\tiny$\pm$.023&1.00\,\tiny$\pm$.001&0.79\,\tiny$\pm$.011 & 0.55\,\tiny$\pm$.024&0.52\,\tiny$\pm$.026&1.00\,\tiny$\pm$.001&0.69\,\tiny$\pm$.012 \\
    AlphaEdit~\cite{fang2025alphaedit} & 0.86\,\tiny$\pm$.014&0.74\,\tiny$\pm$.015&0.95\,\tiny$\pm$.009&0.85\,\tiny$\pm$.007 & 0.86\,\tiny$\pm$.017&0.73\,\tiny$\pm$.018&0.84\,\tiny$\pm$.011&0.81\,\tiny$\pm$.009 & 0.85\,\tiny$\pm$.019&0.72\,\tiny$\pm$.020&0.68\,\tiny$\pm$.012&0.75\,\tiny$\pm$.010 & 0.63\,\tiny$\pm$.021&0.55\,\tiny$\pm$.022&0.30\,\tiny$\pm$.014&0.49\,\tiny$\pm$.011 \\
    MEMOIR~\cite{wang2026memoir} & 0.89\,\tiny$\pm$.009&0.87\,\tiny$\pm$.011&0.97\,\tiny$\pm$.007&0.91\,\tiny$\pm$.005 & 0.84\,\tiny$\pm$.011&0.78\,\tiny$\pm$.014&0.85\,\tiny$\pm$.008&0.82\,\tiny$\pm$.007 & 0.80\,\tiny$\pm$.012&0.74\,\tiny$\pm$.016&0.83\,\tiny$\pm$.009&0.79\,\tiny$\pm$.007 & 0.73\,\tiny$\pm$.014&0.65\,\tiny$\pm$.017&0.80\,\tiny$\pm$.010&0.73\,\tiny$\pm$.008 \\
    ELDER~\cite{elder} & 0.75\,\tiny$\pm$.016&0.69\,\tiny$\pm$.018&0.92\,\tiny$\pm$.010&0.79\,\tiny$\pm$.009 & 0.66\,\tiny$\pm$.019&0.60\,\tiny$\pm$.022&0.89\,\tiny$\pm$.012&0.72\,\tiny$\pm$.010 & 0.57\,\tiny$\pm$.022&0.52\,\tiny$\pm$.025&0.87\,\tiny$\pm$.014&0.65\,\tiny$\pm$.012 & 0.45\,\tiny$\pm$.024&0.41\,\tiny$\pm$.028&0.83\,\tiny$\pm$.016&0.56\,\tiny$\pm$.013 \\
    MELO~\cite{melo} & 0.78\,\tiny$\pm$.014&0.73\,\tiny$\pm$.016&0.95\,\tiny$\pm$.009&0.82\,\tiny$\pm$.008 & 0.70\,\tiny$\pm$.017&0.64\,\tiny$\pm$.019&0.93\,\tiny$\pm$.011&0.76\,\tiny$\pm$.009 & 0.62\,\tiny$\pm$.019&0.56\,\tiny$\pm$.022&0.90\,\tiny$\pm$.012&0.69\,\tiny$\pm$.010 & 0.51\,\tiny$\pm$.021&0.46\,\tiny$\pm$.024&0.87\,\tiny$\pm$.014&0.61\,\tiny$\pm$.012 \\
    AdaLoRA~\cite{adalora} & 0.60\,\tiny$\pm$.023&0.50\,\tiny$\pm$.025&0.18\,\tiny$\pm$.011&0.43\,\tiny$\pm$.012 & 0.01\,\tiny$\pm$.001&0.01\,\tiny$\pm$.001&0.00\,\tiny$\pm$.000&0.01\,\tiny$\pm$.001 & 0.00\,\tiny$\pm$.000&0.00\,\tiny$\pm$.000&0.00\,\tiny$\pm$.000&0.00\,\tiny$\pm$.001 & 0.00\,\tiny$\pm$.000&0.00\,\tiny$\pm$.000&0.00\,\tiny$\pm$.000&0.00\,\tiny$\pm$.001 \\
    \midrule
    LoRA (Exact)~\cite{hu2022lora} & 1.00\,\tiny$\pm$.001&0.99\,\tiny$\pm$.002&1.00\,\tiny$\pm$.001&\textbf{1.00\,\tiny$\pm$.001} & 0.89\,\tiny$\pm$.007&0.95\,\tiny$\pm$.008&1.00\,\tiny$\pm$.001&\textbf{0.95\,\tiny$\pm$.004} & 0.86\,\tiny$\pm$.008&0.93\,\tiny$\pm$.009&1.00\,\tiny$\pm$.001&0.93\,\tiny$\pm$.004 & 0.81\,\tiny$\pm$.009&0.91\,\tiny$\pm$.010&1.00\,\tiny$\pm$.001&\textbf{0.91\,\tiny$\pm$.004} \\
    \textbf{LadderEdit (Ours)} & 1.00\,\tiny$\pm$.001&0.99\,\tiny$\pm$.002&1.00\,\tiny$\pm$.001&\textbf{1.00\,\tiny$\pm$.001} & 0.89\,\tiny$\pm$.006&0.96\,\tiny$\pm$.007&1.00\,\tiny$\pm$.001&\textbf{0.95\,\tiny$\pm$.003} & 0.87\,\tiny$\pm$.006&0.94\,\tiny$\pm$.008&1.00\,\tiny$\pm$.001&\textbf{0.94\,\tiny$\pm$.003} & 0.81\,\tiny$\pm$.007&0.92\,\tiny$\pm$.009&1.00\,\tiny$\pm$.001&\textbf{0.91\,\tiny$\pm$.004} \\
    \midrule
    & \multicolumn{16}{c}{\texttt{\textbf{Qwen2.5-7B}}} \\
    \midrule
    FT & 0.09\,\tiny$\pm$.005&0.08\,\tiny$\pm$.006&0.02\,\tiny$\pm$.001&0.06\,\tiny$\pm$.003 & 0.10\,\tiny$\pm$.006&0.09\,\tiny$\pm$.007&0.01\,\tiny$\pm$.001&0.07\,\tiny$\pm$.003 & 0.08\,\tiny$\pm$.006&0.07\,\tiny$\pm$.008&0.01\,\tiny$\pm$.001&0.05\,\tiny$\pm$.003 & 0.04\,\tiny$\pm$.002&0.03\,\tiny$\pm$.002&0.01\,\tiny$\pm$.001&0.03\,\tiny$\pm$.001 \\
    ROME~\cite{rome} & 0.04\,\tiny$\pm$.002&0.03\,\tiny$\pm$.002&0.01\,\tiny$\pm$.001&0.03\,\tiny$\pm$.001 & 0.01\,\tiny$\pm$.001&0.01\,\tiny$\pm$.001&0.01\,\tiny$\pm$.001&0.01\,\tiny$\pm$.001 & 0.00\,\tiny$\pm$.000&0.00\,\tiny$\pm$.000&0.00\,\tiny$\pm$.000&0.00\,\tiny$\pm$.001 & 0.00\,\tiny$\pm$.000&0.00\,\tiny$\pm$.000&0.00\,\tiny$\pm$.000&0.00\,\tiny$\pm$.001 \\
    MEMIT~\cite{memit} & 0.33\,\tiny$\pm$.002&0.27\,\tiny$\pm$.002&0.30\,\tiny$\pm$.001&0.30\,\tiny$\pm$.001 & 0.05\,\tiny$\pm$.002&0.03\,\tiny$\pm$.002&0.04\,\tiny$\pm$.002&0.04\,\tiny$\pm$.001 & 0.03\,\tiny$\pm$.002&0.01\,\tiny$\pm$.001&0.01\,\tiny$\pm$.001&0.02\,\tiny$\pm$.001 & 0.00\,\tiny$\pm$.000&0.01\,\tiny$\pm$.001&0.01\,\tiny$\pm$.001&0.01\,\tiny$\pm$.001 \\
    GRACE~\cite{grace} & 1.00\,\tiny$\pm$.001&0.36\,\tiny$\pm$.011&1.00\,\tiny$\pm$.001&0.79\,\tiny$\pm$.004 & 1.00\,\tiny$\pm$.001&0.20\,\tiny$\pm$.013&1.00\,\tiny$\pm$.001&0.73\,\tiny$\pm$.004 & 1.00\,\tiny$\pm$.001&0.15\,\tiny$\pm$.015&1.00\,\tiny$\pm$.001&0.72\,\tiny$\pm$.005 & 1.00\,\tiny$\pm$.001&0.04\,\tiny$\pm$.002&1.00\,\tiny$\pm$.001&0.68\,\tiny$\pm$.001 \\
    WISE~\cite{wise} & 0.56\,\tiny$\pm$.017&0.53\,\tiny$\pm$.018&1.00\,\tiny$\pm$.001&0.70\,\tiny$\pm$.008 & 0.61\,\tiny$\pm$.020&0.58\,\tiny$\pm$.022&1.00\,\tiny$\pm$.001&0.73\,\tiny$\pm$.010 & 0.58\,\tiny$\pm$.023&0.55\,\tiny$\pm$.024&0.99\,\tiny$\pm$.002&0.71\,\tiny$\pm$.011 & 0.46\,\tiny$\pm$.025&0.43\,\tiny$\pm$.027&0.99\,\tiny$\pm$.002&0.63\,\tiny$\pm$.012 \\
    AlphaEdit~\cite{fang2025alphaedit} & 0.84\,\tiny$\pm$.014&0.71\,\tiny$\pm$.016&0.92\,\tiny$\pm$.010&0.82\,\tiny$\pm$.008 & 0.81\,\tiny$\pm$.017&0.69\,\tiny$\pm$.019&0.78\,\tiny$\pm$.012&0.76\,\tiny$\pm$.009 & 0.74\,\tiny$\pm$.019&0.66\,\tiny$\pm$.021&0.49\,\tiny$\pm$.013&0.63\,\tiny$\pm$.010 & 0.51\,\tiny$\pm$.022&0.46\,\tiny$\pm$.023&0.18\,\tiny$\pm$.014&0.38\,\tiny$\pm$.012 \\
    MEMOIR~\cite{wang2026memoir} & 0.84\,\tiny$\pm$.010&0.81\,\tiny$\pm$.012&0.93\,\tiny$\pm$.007&0.86\,\tiny$\pm$.006 & 0.78\,\tiny$\pm$.012&0.71\,\tiny$\pm$.014&0.79\,\tiny$\pm$.009&0.76\,\tiny$\pm$.007 & 0.73\,\tiny$\pm$.013&0.67\,\tiny$\pm$.016&0.76\,\tiny$\pm$.010&0.72\,\tiny$\pm$.008 & 0.66\,\tiny$\pm$.014&0.58\,\tiny$\pm$.018&0.72\,\tiny$\pm$.011&0.65\,\tiny$\pm$.008 \\
    ELDER~\cite{elder} & 0.72\,\tiny$\pm$.017&0.66\,\tiny$\pm$.019&0.90\,\tiny$\pm$.011&0.76\,\tiny$\pm$.009 & 0.62\,\tiny$\pm$.020&0.56\,\tiny$\pm$.023&0.87\,\tiny$\pm$.013&0.68\,\tiny$\pm$.011 & 0.53\,\tiny$\pm$.023&0.48\,\tiny$\pm$.026&0.85\,\tiny$\pm$.015&0.62\,\tiny$\pm$.013 & 0.41\,\tiny$\pm$.025&0.37\,\tiny$\pm$.029&0.81\,\tiny$\pm$.016&0.53\,\tiny$\pm$.014 \\
    MELO~\cite{melo} & 0.75\,\tiny$\pm$.014&0.69\,\tiny$\pm$.017&0.93\,\tiny$\pm$.010&0.79\,\tiny$\pm$.008 & 0.66\,\tiny$\pm$.017&0.59\,\tiny$\pm$.020&0.91\,\tiny$\pm$.012&0.72\,\tiny$\pm$.010 & 0.58\,\tiny$\pm$.019&0.51\,\tiny$\pm$.023&0.89\,\tiny$\pm$.013&0.66\,\tiny$\pm$.011 & 0.46\,\tiny$\pm$.022&0.41\,\tiny$\pm$.025&0.85\,\tiny$\pm$.014&0.57\,\tiny$\pm$.012 \\
    AdaLoRA~\cite{adalora} & 0.55\,\tiny$\pm$.024&0.46\,\tiny$\pm$.026&0.15\,\tiny$\pm$.012&0.39\,\tiny$\pm$.012 & 0.00\,\tiny$\pm$.000&0.00\,\tiny$\pm$.000&0.00\,\tiny$\pm$.000&0.00\,\tiny$\pm$.001 & 0.00\,\tiny$\pm$.000&0.00\,\tiny$\pm$.000&0.00\,\tiny$\pm$.000&0.00\,\tiny$\pm$.001 & 0.00\,\tiny$\pm$.000&0.00\,\tiny$\pm$.000&0.00\,\tiny$\pm$.000&0.00\,\tiny$\pm$.001 \\
    \midrule
    LoRA (Exact)~\cite{hu2022lora} & 1.00\,\tiny$\pm$.001&0.99\,\tiny$\pm$.002&1.00\,\tiny$\pm$.001&\textbf{1.00\,\tiny$\pm$.001} & 0.85\,\tiny$\pm$.007&0.94\,\tiny$\pm$.009&1.00\,\tiny$\pm$.001&\textbf{0.93\,\tiny$\pm$.004} & 0.82\,\tiny$\pm$.008&0.91\,\tiny$\pm$.010&1.00\,\tiny$\pm$.001&0.91\,\tiny$\pm$.004 & 0.77\,\tiny$\pm$.009&0.89\,\tiny$\pm$.011&1.00\,\tiny$\pm$.001&\textbf{0.89\,\tiny$\pm$.005} \\
    \textbf{LadderEdit (Ours)} & 1.00\,\tiny$\pm$.001&0.99\,\tiny$\pm$.002&1.00\,\tiny$\pm$.001&\textbf{1.00\,\tiny$\pm$.001} & 0.84\,\tiny$\pm$.006&0.94\,\tiny$\pm$.007&1.00\,\tiny$\pm$.001&\textbf{0.93\,\tiny$\pm$.003} & 0.84\,\tiny$\pm$.006&0.93\,\tiny$\pm$.008&1.00\,\tiny$\pm$.001&\textbf{0.92\,\tiny$\pm$.003} & 0.76\,\tiny$\pm$.007&0.87\,\tiny$\pm$.009&1.00\,\tiny$\pm$.001&0.88\,\tiny$\pm$.004 \\
    \bottomrule
  \end{tabular}}
  \label{tab:zsre-main}
\end{table*}
\begin{table*}[t]
  \caption{Sequential editing results on CF. $T$ denotes the number of edits. Mean $\pm$ standard deviation over three random seeds. The best performance per column is marked in \textbf{bold}.}
  \vspace{3pt}
  \centering
  \resizebox{0.95\linewidth}{!}{%
  \begin{tabular}{l|cc|cc|cc|cc!{\vrule width 0.09em}cc|cc|cc|cc}
  \toprule
   \multirow{2}{*}{\textbf{Method}}  & \multicolumn{8}{c!{\vrule width 0.09em}}{\texttt{\textbf{LLaMA-3-8B}}}  & \multicolumn{8}{c}{\texttt{\textbf{Mistral-7B}}} \\
   \cmidrule(lr){2-9} \cmidrule(lr){10-17}
   & \multicolumn{2}{c|}{$T=100$} & \multicolumn{2}{c|}{$T=500$} & \multicolumn{2}{c|}{$T=1{,}000$} & \multicolumn{2}{c!{\vrule width 0.09em}}{$T=2{,}000$}
   & \multicolumn{2}{c|}{$T=100$} & \multicolumn{2}{c|}{$T=500$} & \multicolumn{2}{c|}{$T=1{,}000$} & \multicolumn{2}{c}{$T=2{,}000$}\\
  \cmidrule{2-17}
    & Rel. & Loc. & Rel. & Loc. & Rel. & Loc. & Rel. & Loc.
    & Rel. & Loc. & Rel. & Loc. & Rel. & Loc. & Rel. & Loc.\\
  \midrule
   FT &0.42\,\tiny$\pm$.004&0.55\,\tiny$\pm$.001&0.25\,\tiny$\pm$.005&0.30\,\tiny$\pm$.001&0.12\,\tiny$\pm$.005&0.12\,\tiny$\pm$.001&0.05\,\tiny$\pm$.002&0.03\,\tiny$\pm$.002&0.39\,\tiny$\pm$.005&0.52\,\tiny$\pm$.001&0.22\,\tiny$\pm$.006&0.27\,\tiny$\pm$.001&0.10\,\tiny$\pm$.006&0.10\,\tiny$\pm$.002&0.03\,\tiny$\pm$.002&0.02\,\tiny$\pm$.001\\
   ROME~\cite{rome} &0.28\,\tiny$\pm$.003&0.45\,\tiny$\pm$.002&0.08\,\tiny$\pm$.004&0.15\,\tiny$\pm$.002&0.02\,\tiny$\pm$.001&0.05\,\tiny$\pm$.002&0.00\,\tiny$\pm$.000&0.02\,\tiny$\pm$.001&0.25\,\tiny$\pm$.003&0.42\,\tiny$\pm$.002&0.06\,\tiny$\pm$.004&0.12\,\tiny$\pm$.003&0.01\,\tiny$\pm$.001&0.04\,\tiny$\pm$.002&0.00\,\tiny$\pm$.000&0.01\,\tiny$\pm$.001\\
   MEMIT~\cite{memit} &0.35\,\tiny$\pm$.002&0.52\,\tiny$\pm$.001&0.12\,\tiny$\pm$.002&0.20\,\tiny$\pm$.001&0.03\,\tiny$\pm$.002&0.08\,\tiny$\pm$.001&0.01\,\tiny$\pm$.001&0.03\,\tiny$\pm$.002&0.31\,\tiny$\pm$.002&0.48\,\tiny$\pm$.001&0.09\,\tiny$\pm$.003&0.16\,\tiny$\pm$.001&0.02\,\tiny$\pm$.001&0.06\,\tiny$\pm$.002&0.00\,\tiny$\pm$.000&0.02\,\tiny$\pm$.001\\
   GRACE~\cite{grace} &0.75\,\tiny$\pm$.001&\textbf{0.96\,\tiny$\pm$.001}&0.71\,\tiny$\pm$.001&\textbf{0.96\,\tiny$\pm$.001}&0.68\,\tiny$\pm$.001&\textbf{0.96\,\tiny$\pm$.001}&0.62\,\tiny$\pm$.001&\textbf{0.95\,\tiny$\pm$.001}&0.72\,\tiny$\pm$.001&\textbf{0.95\,\tiny$\pm$.001}&0.68\,\tiny$\pm$.001&\textbf{0.95\,\tiny$\pm$.001}&0.65\,\tiny$\pm$.001&\textbf{0.95\,\tiny$\pm$.001}&0.59\,\tiny$\pm$.001&\textbf{0.94\,\tiny$\pm$.001}\\
   WISE~\cite{wise} &0.68\,\tiny$\pm$.014&0.88\,\tiny$\pm$.002&0.62\,\tiny$\pm$.017&0.86\,\tiny$\pm$.002&0.55\,\tiny$\pm$.019&0.83\,\tiny$\pm$.003&0.45\,\tiny$\pm$.021&0.77\,\tiny$\pm$.003&0.65\,\tiny$\pm$.016&0.86\,\tiny$\pm$.002&0.58\,\tiny$\pm$.019&0.83\,\tiny$\pm$.003&0.50\,\tiny$\pm$.022&0.80\,\tiny$\pm$.003&0.40\,\tiny$\pm$.024&0.73\,\tiny$\pm$.003\\
   AlphaEdit~\cite{fang2025alphaedit} &0.62\,\tiny$\pm$.012&0.85\,\tiny$\pm$.008&0.58\,\tiny$\pm$.014&0.70\,\tiny$\pm$.010&0.52\,\tiny$\pm$.016&0.48\,\tiny$\pm$.011&0.42\,\tiny$\pm$.018&0.22\,\tiny$\pm$.012&0.58\,\tiny$\pm$.014&0.82\,\tiny$\pm$.009&0.54\,\tiny$\pm$.017&0.66\,\tiny$\pm$.011&0.47\,\tiny$\pm$.019&0.43\,\tiny$\pm$.012&0.37\,\tiny$\pm$.021&0.18\,\tiny$\pm$.014\\
   MEMOIR~\cite{wang2026memoir} &0.78\,\tiny$\pm$.008&0.88\,\tiny$\pm$.006&0.75\,\tiny$\pm$.010&0.94\,\tiny$\pm$.007&0.72\,\tiny$\pm$.011&0.93\,\tiny$\pm$.008&0.66\,\tiny$\pm$.012&0.90\,\tiny$\pm$.009&0.76\,\tiny$\pm$.009&0.86\,\tiny$\pm$.007&0.72\,\tiny$\pm$.011&0.93\,\tiny$\pm$.008&0.69\,\tiny$\pm$.012&0.91\,\tiny$\pm$.009&0.63\,\tiny$\pm$.014&0.88\,\tiny$\pm$.010\\
   ELDER~\cite{elder} &0.52\,\tiny$\pm$.014&0.82\,\tiny$\pm$.009&0.46\,\tiny$\pm$.017&0.78\,\tiny$\pm$.011&0.39\,\tiny$\pm$.019&0.73\,\tiny$\pm$.012&0.30\,\tiny$\pm$.021&0.66\,\tiny$\pm$.013&0.49\,\tiny$\pm$.016&0.80\,\tiny$\pm$.010&0.43\,\tiny$\pm$.019&0.76\,\tiny$\pm$.012&0.36\,\tiny$\pm$.022&0.71\,\tiny$\pm$.014&0.27\,\tiny$\pm$.024&0.64\,\tiny$\pm$.016\\
   MELO~\cite{melo} &0.56\,\tiny$\pm$.012&0.86\,\tiny$\pm$.008&0.50\,\tiny$\pm$.014&0.82\,\tiny$\pm$.010&0.43\,\tiny$\pm$.016&0.78\,\tiny$\pm$.011&0.34\,\tiny$\pm$.018&0.71\,\tiny$\pm$.012&0.53\,\tiny$\pm$.014&0.84\,\tiny$\pm$.009&0.47\,\tiny$\pm$.017&0.80\,\tiny$\pm$.011&0.40\,\tiny$\pm$.019&0.76\,\tiny$\pm$.012&0.31\,\tiny$\pm$.021&0.69\,\tiny$\pm$.014\\
   AdaLoRA~\cite{adalora} &0.18\,\tiny$\pm$.020&0.32\,\tiny$\pm$.010&0.02\,\tiny$\pm$.001&0.02\,\tiny$\pm$.001&0.00\,\tiny$\pm$.000&0.01\,\tiny$\pm$.001&0.00\,\tiny$\pm$.000&0.00\,\tiny$\pm$.000&0.15\,\tiny$\pm$.023&0.30\,\tiny$\pm$.011&0.01\,\tiny$\pm$.001&0.02\,\tiny$\pm$.001&0.00\,\tiny$\pm$.000&0.01\,\tiny$\pm$.001&0.00\,\tiny$\pm$.000&0.00\,\tiny$\pm$.000\\
  \midrule
   LoRA (Exact)~\cite{hu2022lora} &0.82\,\tiny$\pm$.005&0.85\,\tiny$\pm$.002&\textbf{0.78\,\tiny$\pm$.006}&\textbf{0.96\,\tiny$\pm$.002}&0.75\,\tiny$\pm$.007&0.95\,\tiny$\pm$.003&0.71\,\tiny$\pm$.007&0.93\,\tiny$\pm$.003&0.80\,\tiny$\pm$.006&0.83\,\tiny$\pm$.002&0.76\,\tiny$\pm$.007&\textbf{0.95\,\tiny$\pm$.003}&0.72\,\tiny$\pm$.008&0.93\,\tiny$\pm$.003&0.68\,\tiny$\pm$.009&0.91\,\tiny$\pm$.003\\
   \textbf{LadderEdit (Ours)} &\textbf{0.83\,\tiny$\pm$.004}&0.90\,\tiny$\pm$.002&\textbf{0.78\,\tiny$\pm$.005}&\textbf{0.96\,\tiny$\pm$.002}&\textbf{0.77\,\tiny$\pm$.005}&\textbf{0.96\,\tiny$\pm$.003}&\textbf{0.74\,\tiny$\pm$.006}&0.94\,\tiny$\pm$.003&\textbf{0.81\,\tiny$\pm$.005}&0.88\,\tiny$\pm$.002&\textbf{0.77\,\tiny$\pm$.006}&\textbf{0.95\,\tiny$\pm$.003}&\textbf{0.74\,\tiny$\pm$.006}&0.94\,\tiny$\pm$.003&\textbf{0.71\,\tiny$\pm$.007}&0.92\,\tiny$\pm$.003\\
  \bottomrule
  \end{tabular}}
  \label{tab:counterfact-main}
\end{table*}
\paragraph{Datasets.}
We conduct experiments on three benchmarks: ZsRE~\cite{levy2017zero}, CounterFact (CF)~\cite{rome,memit}, and WikiBigEdit~\cite{thede2025wikibigedit}. ZsRE is a question-answering dataset for evaluating factual knowledge and generalization in language models. CF tests the ability to edit and preserve specific factual associations in a controlled setting. WikiBigEdit is a large-scale benchmark for long-horizon, up to 50{,}000 edits, sequential editing.

We evaluate LadderEdit on LLaMA-3-8B~\cite{LLaMA3}, Mistral-7B~\cite{jiang2023mistral}, and Qwen2.5-7B~\cite{yang2025qwen3}. We compare against representative direct-editing methods, memory- and routing-based lifelong editors, constrained-update methods, LoRA/MoE-style edit-memory baselines, adaptive-rank LoRA, and the full-resolution exact LoRA reference. Detailed baseline descriptions are deferred to Appendix~\ref{app:appendix-baselines}.

\paragraph{Edit lookup protocol.}
LadderEdit evaluates storage for edit-local memories, not edit selection. For edit-local methods, each rewrite, generalization, and locality probe is associated with the edit that generated it. Exact LoRA and LadderEdit use the same association, so they differ only in the stored representation returned after lookup. Methods whose routing is part of the original algorithm, such as memory- or MoE-based lifelong editors, are evaluated with their native routing mechanisms.

We report reliability, generalization, locality, and their average when applicable. Locality is the fraction of unrelated probes whose prediction remains unchanged relative to the frozen base model. 1.00 means no failures on the finite locality probe set.

\paragraph{Probe split protocol.}
For each edit, the benchmark-provided prompts are partitioned into
three disjoint sets at edit-construction time. The \emph{audit
set} contains the rewrite prompt and a held-out subset of
rephrase/generalization variants, used only at insertion time to
score $S_i(r)$ and gate the contract gap $H_i^{(r)}$. The
\emph{validation set} is a separate stream used only to select
audit thresholds, which are then frozen before the test stream is
touched. The \emph{test set} contains the remaining
rephrase/generalization variants and unrelated locality probes;
it is used only to report Reliability, Generalization, and
Locality in the main tables and is never seen by the audit.
Audit and test sets share neither prompts nor paraphrases of the
same prompt. Full split statistics, audit-test correlation, and
a leaky-versus-clean comparison are reported in
Appendix~\ref{app:probe-split}. More details in Appendix~\ref{app:appendix-training-seeds}.

\subsection{Results}

\paragraph{Main Results.}
Tables~\ref{tab:zsre-main} and~\ref{tab:counterfact-main} show that LadderEdit preserves the plasticity of edit-local adapters while reducing their persistent storage. On ZsRE, FT, ROME, and MEMIT degrade quickly, and GRACE preserves locality but fails to generalize, especially on Mistral-7B and Qwen2.5-7B. Exact LoRA remains the strongest full-storage reference because each edit has isolated capacity, but LadderEdit tracks it after compression: on LLaMA-3-8B it improves the average over exact LoRA at $T=500$ (0.97 vs.\ 0.93) and matches it at $T=1{,}000$ and $T=2{,}000$; on Mistral-7B it slightly improves reliability or generalization at longer horizons; on Qwen2.5-7B it stays within 0.01 to 0.02 average of exact LoRA at the largest horizons. CounterFact gives the same message in a setting that emphasizes locality: at $T=2{,}000$, LadderEdit moves exact LoRA from 0.71/0.93 to 0.74/0.94 in reliability/locality on LLaMA-3-8B and from 0.68/0.91 to 0.71/0.92 on Mistral-7B. These gains suggest that compression is not merely lossy: low-rank sketches can remove overly specific directions, while the audit promotes edits to higher rank when more fidelity is needed.

\begin{figure}[t]
    \centering
    \includegraphics[width=\linewidth]{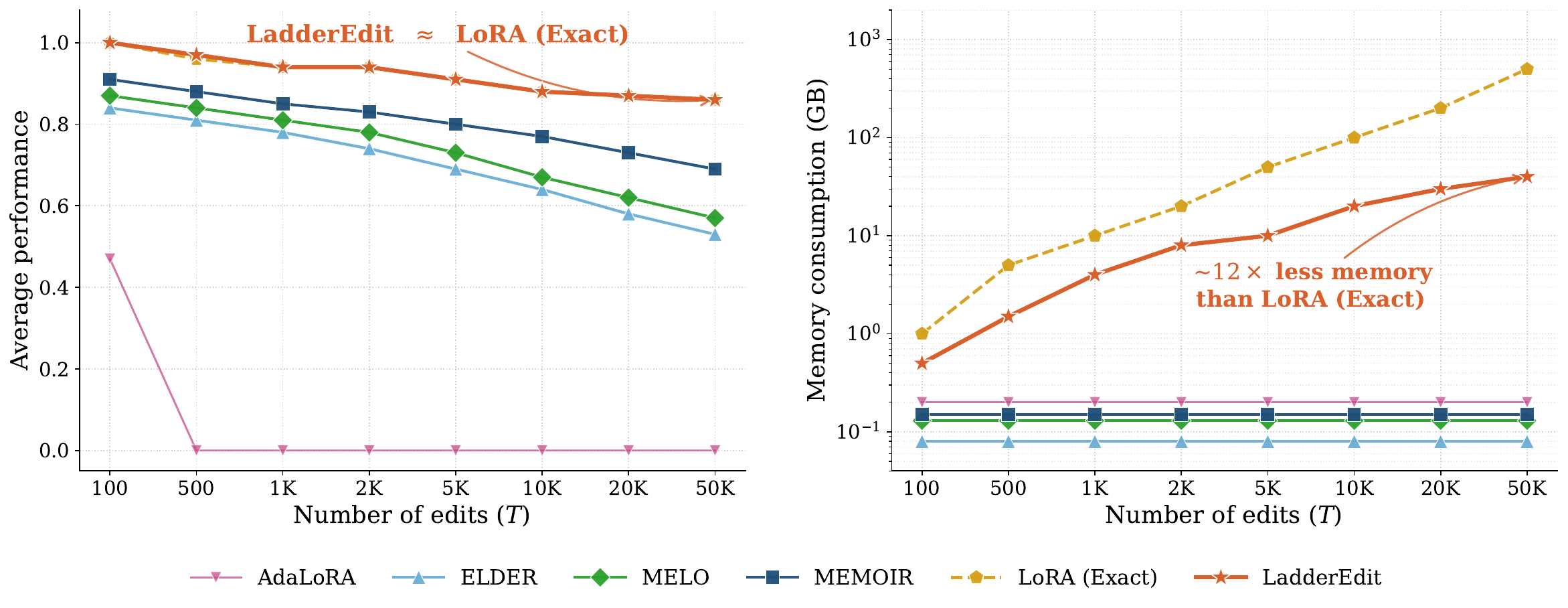}
    \caption{Scaling reference on WikiBigEdit: average performance against persistent edit memory.}
    \label{fig:scaling-perf-memory}
\end{figure}

\paragraph{WikiBigEdit Long-Horizon Scaling.}
At $T=50{,}000$ on WikiBigEdit
(Table~\ref{tab:wikibigedit-scaling}), LadderEdit reaches 0.75
reliability, 0.84 generalization, 0.98 locality, and 0.86 average.
GRACE holds reliability and locality but drops generalization to
0.28; WISE loses both reliability and generalization; MEMOIR reaches
only 0.69 average. Together with the memory trade-off in
Figure~\ref{fig:scaling-perf-memory}, LadderEdit preserves
edit-local behavior at this scale while substantially reducing
persistent storage.

\begin{table}[t]
  \centering
  \small
  \caption{Editing performance on Qwen2.5-7B at long-horizon scale on WikiBigEdit. Mean $\pm$ standard deviation over three random seeds.}
  \label{tab:wikibigedit-scaling}
  \resizebox{\linewidth}{!}{%
    \begin{tabular}{l|cccc|cccc}
      \toprule
      \multirow{2}{*}{\textbf{Method}}
        & \multicolumn{4}{c|}{$T={10{,}000}$}
        & \multicolumn{4}{c}{$T={50{,}000}$} \\
      \cmidrule(lr){2-5}\cmidrule(lr){6-9}
        & Rel. & Gen. & Loc. & Avg.
        & Rel. & Gen. & Loc. & Avg. \\
      \midrule
      MEMIT~\cite{memit} & 0.00\,\tiny$\pm$.000 & 0.00\,\tiny$\pm$.000 & 0.00\,\tiny$\pm$.000 & 0.00\,\tiny$\pm$.001 & N/A & N/A & N/A & N/A \\
      GRACE~\cite{grace} & 1.00\,\tiny$\pm$.001 & 0.31\,\tiny$\pm$.017 & 1.00\,\tiny$\pm$.001 & 0.77\,\tiny$\pm$.006 & 0.99\,\tiny$\pm$.002 & 0.28\,\tiny$\pm$.019 & 1.00\,\tiny$\pm$.001 & 0.76\,\tiny$\pm$.006 \\
      WISE~\cite{wise} & 0.59\,\tiny$\pm$.027 & 0.57\,\tiny$\pm$.029 & 1.00\,\tiny$\pm$.001 & 0.72\,\tiny$\pm$.013 & 0.43\,\tiny$\pm$.030 & 0.41\,\tiny$\pm$.032 & 1.00\,\tiny$\pm$.001 & 0.61\,\tiny$\pm$.015 \\
      AlphaEdit~\cite{fang2025alphaedit} & 0.72\,\tiny$\pm$.023 & 0.77\,\tiny$\pm$.025 & 0.74\,\tiny$\pm$.015 & 0.74\,\tiny$\pm$.012 & 0.62\,\tiny$\pm$.026 & 0.60\,\tiny$\pm$.028 & 0.71\,\tiny$\pm$.017 & 0.64\,\tiny$\pm$.014 \\
      MEMOIR~\cite{wang2026memoir} & 0.72\,\tiny$\pm$.015 & 0.75\,\tiny$\pm$.019 & 0.83\,\tiny$\pm$.012 & 0.77\,\tiny$\pm$.009 & 0.66\,\tiny$\pm$.017 & 0.68\,\tiny$\pm$.022 & 0.73\,\tiny$\pm$.013 & 0.69\,\tiny$\pm$.010 \\
      ELDER~\cite{elder} & 0.54\,\tiny$\pm$.027 & 0.49\,\tiny$\pm$.031 & 0.88\,\tiny$\pm$.017 & 0.64\,\tiny$\pm$.015 & 0.41\,\tiny$\pm$.030 & 0.36\,\tiny$\pm$.035 & 0.82\,\tiny$\pm$.019 & 0.53\,\tiny$\pm$.017 \\
      MELO~\cite{melo} & 0.58\,\tiny$\pm$.023 & 0.53\,\tiny$\pm$.027 & 0.91\,\tiny$\pm$.015 & 0.67\,\tiny$\pm$.013 & 0.45\,\tiny$\pm$.026 & 0.40\,\tiny$\pm$.030 & 0.85\,\tiny$\pm$.017 & 0.57\,\tiny$\pm$.014 \\
      AdaLoRA~\cite{adalora} & 0.00\,\tiny$\pm$.000 & 0.00\,\tiny$\pm$.000 & 0.00\,\tiny$\pm$.000 & 0.00\,\tiny$\pm$.001 & 0.00\,\tiny$\pm$.000 & 0.00\,\tiny$\pm$.000 & 0.00\,\tiny$\pm$.000 & 0.00\,\tiny$\pm$.001 \\
      \midrule
      \textbf{LadderEdit (Ours)} & 0.77\,\tiny$\pm$.008 & 0.87\,\tiny$\pm$.010 & 0.99\,\tiny$\pm$.002 & \textbf{0.88\,\tiny$\pm$.004} & 0.75\,\tiny$\pm$.009 & 0.84\,\tiny$\pm$.011 & 0.98\,\tiny$\pm$.004 & \textbf{0.86\,\tiny$\pm$.005} \\
      \bottomrule
    \end{tabular}}
\end{table}
 
\paragraph{Memory Efficiency and Ablation.}
The memory results support LadderEdit's central claim: lifelong
edit storage should be treated as per-edit resolution allocation.
Table~\ref{tab:heterogeneity} reports the empirical population
structure of the edit stream and confirms the heterogeneity
assumed in Section~\ref{sec:problem-setting}. On both ZsRE and
CounterFact, roughly 80\% of edits are sketch-sufficient at
rank~1, a 13 to 17\% tail requires rank~2 or higher to satisfy
the behavioral contract, and a smaller 3 to 5\% subset requires
the full acquisition rank. A 4 to 7\% compression-regularized
population, where the rank-1 sketch passes the contract that
exact LoRA fails, shows that the full update is not a behavioral
ceiling. This distribution justifies the ladder design: a single
fixed rank cannot serve all edits, but a small higher-rank
budget plus a cheap default rung suffices for the bulk of the
stream. Per-rank compression statistics (memory, retained energy,
false retirements) are reported in
Appendix~\ref{app:fixed-rank-tier}. A predictor ablation isolating the spectral score, the
behavioral margin, and their product against random and oracle
baselines is reported in
Appendix~\ref{app:predictor-ablation}; the combined predictor
approaches oracle utility at far fewer audits per edit than any
component alone.

Table~\ref{tab:frontier-diagnostics} tests the memory primitive
directly. The exact-cache baseline is validation-selected: under
a budget of $K$ full edits, it keeps those with the largest
sketch-only contract gap and drops the rest. At $5.57\times$ (CF)
and $5.63\times$ (ZsRE) compression, it retains about 91 of 512
edits, reaching utility 0.434 on CF and 0.420 on ZsRE; dropped
edits have no representation. LadderEdit covers every edit with a
rank-1 sketch: at $8\times$ compression, the sketch alone reaches
0.753 (CF) and 0.765 (ZsRE), exceeding exact-cache by roughly
0.33 at lower memory, so coverage drives most of the gap.
Behavioral rung selection adds the rest. At the matched budget,
a static mixed-rank policy reaches 0.815 (CF) and 0.796 (ZsRE);
the dual-budget ladder, spending that budget selectively on
audit-flagged edits, reaches 0.862 (CF) and 0.866 (ZsRE). The
0.05 gap between static and dual-budget at identical memory is
the signature of audit-driven allocation. Exact-cache fails by
absence; LadderEdit degrades by resolution.

Finally, Table~\ref{tab:cost-decomposition} shows that this frontier translates into deployed storage savings. LadderEdit is exact-first, so it does not reduce the initial LoRA acquisition cost; its benefit is retained edit memory after acquisition. At $T=10{,}000$ on LLaMA-3-8B, the audit increases per-edit insertion time relative to exact LoRA, but persistent memory drops from 100.93 GB to 19.44 GB, a $5.19\times$ reduction. Because exact LoRA and LadderEdit use the same edit-selection assumption, this is a storage-layer gain: the lookup mechanism is unchanged, and LadderEdit changes only the representation returned after lookup. Additional qualitative examples in Appendix~\ref{app:appendix-examples} and stress tests in Appendix~\ref{app:appendix-stress} show the same mechanism at the prompt and robustness levels: the edit stream is heterogeneous, many edits are sketch-sufficient, and a smaller subset needs higher rank to satisfy the measured contract.

\begin{table}[t]
\centering
\small
\caption{Population structure of edit streams at $N=2{,}000$. Categories are determined by the behavioral audit on audit-split
probes. Most edits are over-resolved by the full LoRA update.}
\resizebox{0.8\linewidth}{!}{%
\begin{tabular}{lcc}
\toprule
Diagnostic case & ZsRE & CF \\
\midrule
Sketch-sufficient (rank-1 passes) & 81.2\% & 76.4\% \\
Detail-needed (rank $\geq 2$ required) & 13.4\% & 16.8\% \\
Residual-needed (rank $R_{\max}$ required) & 3.1\% & 4.7\% \\
Compression-regularized (sketch $>$ exact) & 4.3\% & 6.7\% \\
\bottomrule
\end{tabular}%
}
\label{tab:heterogeneity}
\end{table}

\begin{table}[t]
  \centering
  \small
  \caption{Budget-matched strict-memory frontier at $N=512$. LadderEdit preserves coverage and allocates residual fidelity through behavioral rung selection.}
  \label{tab:frontier-diagnostics}
  \resizebox{\linewidth}{!}{%
  \begin{tabular}{llrrr}
    \toprule
    Dataset & Policy & Memory & Comp. & Utility \\
    \midrule
    CF   & exact-cache (matched)      & 0.180 & 5.57 & 0.434 \\
    CF   & rank-1 sketch only         & 0.125 & 8.00 & 0.753 \\
    CF   & static mixed-rank          & 0.180 & 5.57 & 0.815 \\
    CF   & dual-budget LadderEdit     & 0.180 & 5.57 & 0.862 \\
    \midrule
    ZsRE & exact-cache (matched)      & 0.178 & 5.63 & 0.420 \\
    ZsRE & rank-1 sketch only         & 0.124 & 8.09 & 0.765 \\
    ZsRE & static mixed-rank          & 0.179 & 5.58 & 0.796 \\
    ZsRE & dual-budget LadderEdit     & 0.179 & 5.58 & 0.866 \\
    \bottomrule
  \end{tabular}}
\end{table}

\begin{figure}[t]
    \centering
    \begin{subfigure}{0.48\linewidth}
        \includegraphics[width=\linewidth]{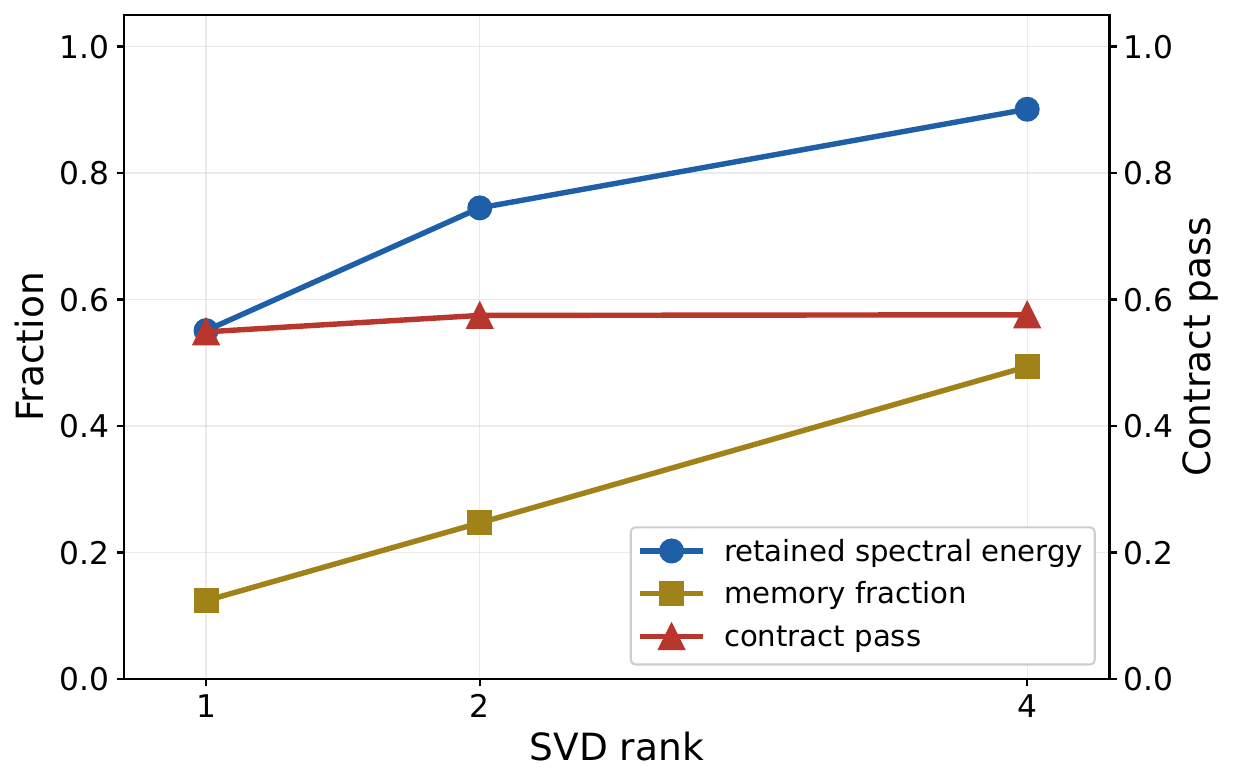}
        \caption{ZsRE}
    \end{subfigure}
    \hfill
    \begin{subfigure}{0.48\linewidth}
        \includegraphics[width=\linewidth]{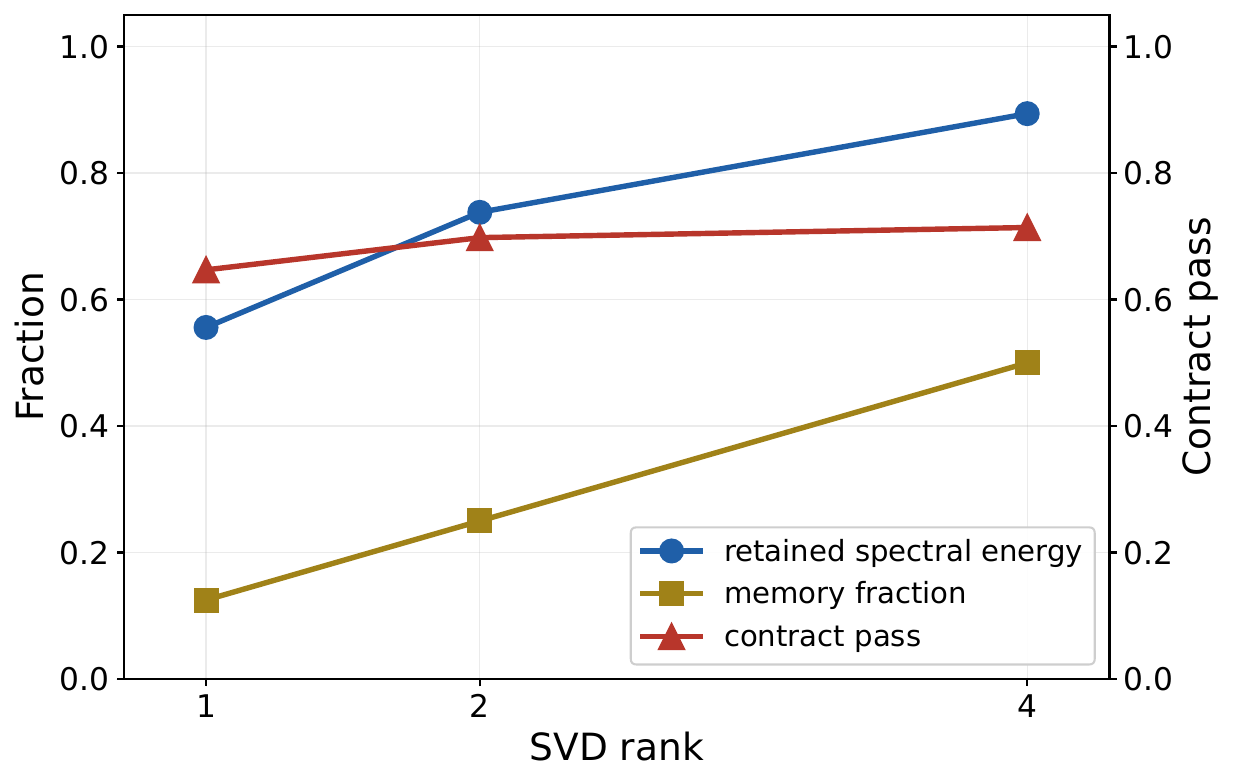}
        \caption{CF}
    \end{subfigure}
    \caption{Spectral-filter frontier analysis on ZsRE and CF at $N=512$ edits. The plots show the tradeoff between memory compression and behavioral contract utility for different rank tiers and allocation strategies.}
    \vspace{3pt}
    \label{fig:spectral-frontier}
\end{figure}

\begin{table}[!t]
  \centering
  \small
  \caption{Cost decomposition at $T=10{,}000$ on LLaMA-3-8B. LadderEdit reduces this storage by $5.19\times$ relative to exact LoRA. Serving path summarizes the additional edit mechanism used at inference.}
  \label{tab:cost-decomposition}
  \resizebox{\linewidth}{!}{%
  \begin{tabular}{lcccl}
    \toprule
    Method & Time & Peak Mem. & Persistent & Serving path \\
    & (ms) & (GB) & (GB) & \\
    \midrule
    LoRA (Exact)~\cite{hu2022lora} & 86  & 18.4 & 100.93 & exact-adapter lookup \\
    MELO~\cite{melo}               & 91  & 18.6 & 100.93 & retrieval \\
    ELDER~\cite{elder}             & 95  & 19.1 & 0.08   & MoE routing \\
    MEMOIR~\cite{wang2026memoir}   & 104 & 17.9 & 0.14   & mask gate \\
    \midrule
    \textbf{LadderEdit}            & 124 & 18.8 & 19.44  & ladder lookup \\
    \bottomrule
  \end{tabular}}
\end{table}

\section{Conclusion}
\label{sec:conclusion}

We introduced LadderEdit, a residual-ladder storage
controller for lifelong editing. LadderEdit first uses a spectral predictor to select a
candidate rank for each acquired edit, and a behavioral audit
verifies the proposal and promotes the edit along the ladder only
when the performance of that edit demands it. Across ZsRE, CF, and
WikiBigEdit on LLaMA-3-8B, Mistral-7B, and Qwen2.5-7B, LadderEdit
tracks exact LoRA while reducing persistent memory by $5.2\times$
and remaining effective at 50{,}000 edits. These results show that
most exact adapters are over-resolved: low-rank sketches preserve
the edit contract for the bulk of the stream, while higher
resolution is needed only for a smaller hard edits subset. The key question in lifelong edit storage is not only \emph{which} edits to keep, but also \emph{how much} of each edit should be retained.

\section{Limitations}
LadderEdit is not constant-memory: every edit retains at least a
rank-1 sketch, so persistent storage grows linearly with the
stream, and future work could explore sketch downgrading or
merging to bound the total. The behavioral audit adds
insertion-time latency (124~ms versus 86~ms at $T{=}10{,}000$ on
LLaMA-3-8B), which could be reduced by batching audits or running
them asynchronously. The audit also depends on probe coverage and
does not guarantee correctness outside the probed set, motivating
richer probe construction or learned probe selection. LadderEdit
is a storage-layer method and inherits the retrieval mechanism
used by the full-resolution memory bank; integration with broader
production retrieval stacks remains open. The spectral score that
proposes candidate ranks is a calibrated heuristic, and replacing
it with a predictor trained on audit outcomes is a natural
extension. Finally, our experiments span three benchmarks and
three 7B and 8B backbones, with long-horizon scaling shown to
$50{,}000$ edits; validating LadderEdit at larger scales and with
non-LoRA edit writers remains future work.

\paragraph{Potential Risks.}
LadderEdit reduces the persistent cost of injecting and retaining
edits in a large language model, which could lower the operational
barrier to large-scale knowledge modification. We see three
concrete risks.
First, the same machinery that enables efficient correction of
outdated or harmful facts can be used to insert false or
adversarial edits at scale, since the audit verifies behavioral
contract satisfaction rather than factual correctness of the
target. Mitigations include provenance tracking on the edit
stream, signed edit acquisition, and downstream factuality
checking, none of which are part of the method itself.
Second, the behavioral audit relies on a probe bank whose coverage
is finite; edits whose effects fall outside the probed semantic
neighborhood may pass the audit while exhibiting unintended
behavior on unprobed prompts. This is a soft-coverage rather than
a hard-correctness guarantee, and we emphasize in the appendix
that the audit should not be treated as a safety certificate.
Third, the residual-ladder representation makes individual edits
less inspectable than uncompressed adapters: a low-rank sketch can be
behaviorally equivalent to an exact LoRA update on the probed contract
while retaining different parameter directions. In safety-critical deployments, this argues for keeping edit-level audit logs and
periodic reaudits against the exact update rather than relying on stored
representations alone. 
We do not release any new model weights or edit corpora and rely
on public benchmarks; the method's broader societal effect depends
on the editing workflow it is deployed in rather than on the
storage primitive itself.

\section{Ethics Considerations}

Model editing can be used to correct errors and update knowledge, but it can also be misused to insert false or harmful facts. Edits require provenance, authorization, validation, and auditing to ensure responsible deployment. LadderEdit is a memory allocation mechanism for model editing; it does not serve as a truth or safety filter.

\section*{Acknowledgment}
This study was supported in part by NIH R21EB034911, Gemini Academic Program Award, and NVIDIA Academic Grant Program.

\bibliography{custom}

\clearpage
\appendix \section*{Appendix} \startcontents[sections] \printcontents[sections]{l}{1}{\setcounter{tocdepth}{2}}


\section{Composing LadderEdit with Learned Retrieval}
\label{app:appendix_retrieval}

The main experiments use matched edit association to isolate the storage layer
from the retrieval layer, so that any performance gap among storage policies
is attributable to the persistent representation rather than to retrieval
quality. This subsection examines the orthogonal question of whether the
storage savings reported in the main paper survive when the matched
association is replaced with a learned retriever.

We evaluate three retrieval strategies on LLaMA-3-8B / ZsRE at $T=1{,}000$: a learned Contriever-style dual encoder trained on edit prompts, a learned E5-large retriever, and a frozen off-the-shelf MPNet embedding (no retrieval training). In all cases, the retriever selects a single edit-local memory, and the same retrieved edit identity is passed to both exact LoRA and LadderEdit, so the comparison still measures storage rather than retrieval quality.

Table~\ref{tab:retrieval-cell} reports sensitivity to retrieval on
ZsRE / LLaMA-3-8B at $T=1{,}000$.

\begin{table}[h]
\centering
\small
\resizebox{\linewidth}{!}{%
\begin{tabular}{lccccc}
\toprule
Setting & Rel.\ & Gen.\ & Loc.\ & Avg.\ & $\Delta$ vs. matched \\
\midrule
Matched association (Exact LoRA) & 0.95 & 0.91 & 1.00 & 0.95 & --- \\
Matched association (LadderEdit) & 0.95 & 0.91 & 1.00 & 0.95 & --- \\
Contriever retriever (Exact LoRA) & 0.91 & 0.86 & 0.99 & 0.92 & $-0.03$ \\
Contriever retriever (LadderEdit) & 0.90 & 0.86 & 0.99 & 0.92 & $-0.03$ \\
E5-large retriever (LadderEdit) & 0.88 & 0.85 & 0.99 & 0.91 & $-0.04$ \\
MPNet (frozen) (LadderEdit) & 0.85 & 0.81 & 0.98 & 0.88 & $-0.07$ \\
\midrule
Persistent memory ratio & --- & --- & --- & --- & $5.2\times$ preserved \\
\bottomrule
\end{tabular}%
}
\caption{Retrieval sensitivity on ZsRE / LLaMA-3-8B at $T=1{,}000$. The persistent-memory ratio is preserved under all retrieval settings in the table because the retriever is shared between exact LoRA and LadderEdit. Locality is barely affected (0.99 $\rightarrow$ 0.98 under MPNet) because retrieved adapters still land on real edit directions rather than random directions.}
\label{tab:retrieval-cell}
\end{table}

Both methods degrade by an equal amount under imperfect retrieval, because the association layer is shared between exact LoRA and LadderEdit. The relative ordering of storage methods is therefore preserved under all retrieval strategies in Table~\ref{tab:retrieval-cell}.

\subsection{Edit-Stream Heterogeneity }
\label{app:heterogeneity}
Table~\ref{tab:heterogeneity} of the main text reports the
overall population structure of the edit stream. Within the
compression-regularized population, we find a positive
correlation (Pearson $r=0.38$ on ZsRE, $r=0.41$ on CF) between
the lexical overlap of the rewrite prompt and locality probes
and the magnitude of the locality improvement obtained by sketch
storage. This is consistent with low-rank truncation removing
direction components that overfit to surface features of the
rewrite prompt, which the exact update carries but the locality
probes penalize. Section~\ref{app:rung-allocation} reports the
corresponding ladder allocation decisions.

\subsection{Comparison with Static Mixed-Rank Allocation}
\label{app:static-mixed-rank}

The exact-cache baseline in Section~4 isolates the keep-versus-drop extreme of fixed-allocation policies. A second non-adaptive baseline, closer to LadderEdit in absolute memory and not dropping edits, assigns every edit a fixed rank with no behavioral audit. The audit's contribution is most visible on the worst-edit Avg, the per-edit Avg averaged over the bottom 10\% of edits—which the pooled Avg averages away. Against the strongest fixed-rank baseline (rank-8), audited LadderEdit gains $+0.05$ on pooled Avg but $+0.12$ on the worst-edit metric. An oracle controller that selects the per-edit optimal rank reaches 0.96 pooled and 0.78 worst-edit, indicating LadderEdit is within 0.01 pooled and 0.04 worst-edit of the oracle. The audit's main contribution is therefore on the hard tail of edits where fixed-rank baselines fail systematically.

\begin{table}[h]
\centering
\small
\resizebox{\linewidth}{!}{%
\begin{tabular}{lccc}
\toprule
Policy & Memory (norm.) & Pooled Avg.\ & Worst-edit Avg.\ \\
\midrule
Fixed rank-1  & 0.125 & 0.85 (0.008) & --- \\
Fixed rank-2  & 0.250 & 0.89 (0.007) & --- \\
Fixed rank-4  & 0.500 & 0.90 (0.008) & --- \\
Fixed rank-8  & 1.000 & 0.90 (0.007) & 0.62 \\
Uniform random rank & 0.562 & 0.87 (0.010) & --- \\
\midrule
LadderEdit (audited) & 0.19 & 0.95 (0.005) & 0.74 \\
Oracle rank controller & --- & 0.96 & 0.78 \\
\bottomrule
\end{tabular}%
}
\caption{Audit contribution against fixed-rank storage at $T=2{,}000$ on ZsRE / LLaMA-3-8B across three seeds. Memory is normalized to exact LoRA. Pooled Avg.\ is the pooled evaluation Avg.\ across edits; worst-edit Avg.\ is the per-edit Avg.\ averaged over only the bottom 10\% of edits. Standard deviations over three seeds are reported in parentheses when available.}
\label{tab:static-rank}
\end{table}

\paragraph{Trade-offs with the strongest baselines.} GRACE achieves 1.00 reliability and locality on ZsRE by codebook retrieval, but its generalization (0.34 on LLaMA-3 at $T=2{,}000$) means it cannot answer paraphrased queries. MEMOIR achieves 0.69 Avg.\ at $T=10{,}000$ in only 0.14~GB of persistent memory, a $138\times$ memory advantage over LadderEdit (19.4~GB). LadderEdit's 0.11 Avg.\ improvement over MEMOIR at $T=10{,}000$ is therefore an explicit memory-for-fidelity trade, not a strict dominance. Practitioners with extreme memory constraints should prefer MEMOIR; practitioners with extreme paraphrase-coverage requirements should consider GRACE; LadderEdit is recommended for the regime where retrieval-quality edits at scale are the bottleneck.

\subsection{Audit Score Weights}
\label{app:audit-weights}
 
The audit score in Eq.~\ref{eq:contract-gap} combines reliability, generalization, and locality residuals as a weighted sum, $s(z) = w_R \cdot \Delta_R + w_G \cdot \Delta_G + w_L \cdot \Delta_L$, with default weights $w_R = 1.0$, $w_G = 0.8$, and $w_L = 1.2$. The asymmetry between $w_L$ and the other two weights is deliberate. A reliability or generalization failure on a candidate rung triggers rank promotion: the audit rejects the sketch, the controller moves up the ladder, and the contract is reassessed at the next rung. The failure is recoverable within the same insertion. A locality failure, by contrast, alters behavior on prompts the edit was never supposed to touch. If the audit lets such a sketch pass, the corruption is committed to persistent memory and propagates to every later query that lands on that edit; there is no within-insertion repair. We therefore weight locality higher than the recoverable metrics, treating $w_L > w_R, w_G$ as a soft asymmetry between catastrophic and recoverable failure modes.
 
\begin{figure}[t]
  \centering
  \includegraphics[width=\linewidth]{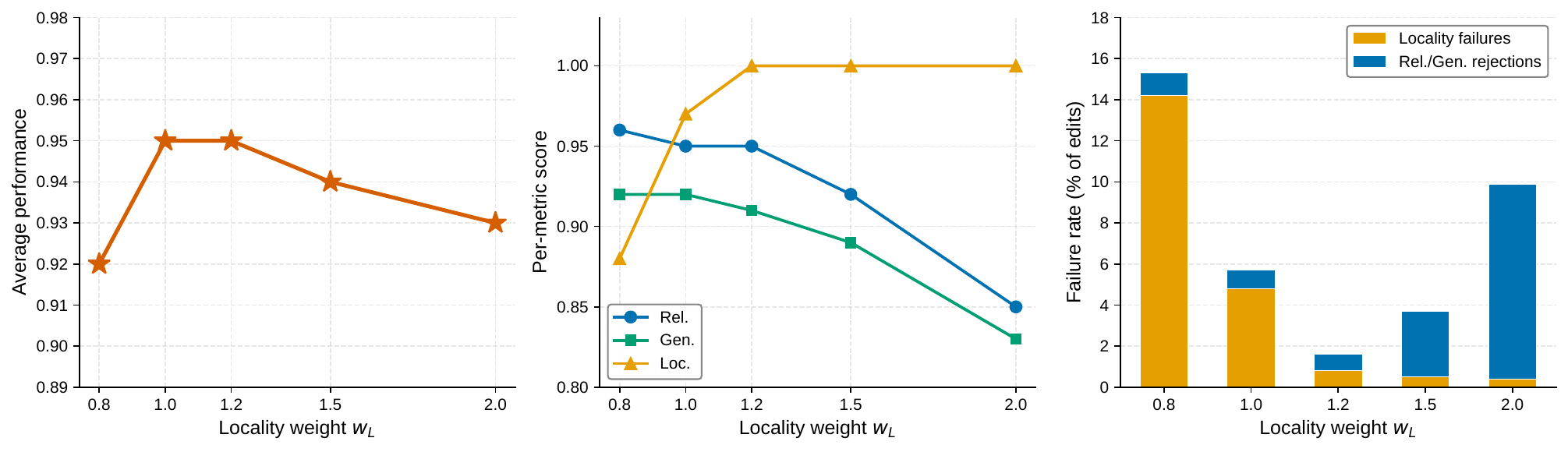}
  \caption{Sensitivity of LadderEdit to the locality weight $w_L$ on ZsRE / LLaMA-3-8B at $T=1{,}000$, with $w_R$ and $w_G$ held fixed. Left: average performance, peaking at the default $w_L = 1.2$ with a stable region across $[1.0, 1.5]$. Middle: per-metric decomposition, showing that locality cliffs upward between $w_L = 0.8$ and $w_L = 1.0$ while reliability and generalization degrade for $w_L \geq 1.5$. Right: failure-mode decomposition, with locality failures dominating at low $w_L$ and reliability or generalization rejections dominating at high $w_L$; the default minimizes the total failure rate.}
  \label{fig:audit-weights-sweep}
\end{figure}

The default $w_L = 1.2$ is selected by a sweep on the validation split. We evaluate $w_L \in \{0.8, 1.0, 1.2, 1.5, 2.0\}$ holding $w_R$ and $w_G$ fixed; Figure~\ref{fig:audit-weights-sweep} reports the result on ZsRE / LLaMA-3-8B at $T=1{,}000$. The choice is not knife-edge: average performance stays within $0.01$ of the peak for $w_L \in [1.0, 1.5]$, and drops by $0.02$ at $w_L = 0.8$ and $0.03$ at $w_L = 2.0$. The default sits at the midpoint of this stable region.
 
The per-metric decomposition (Figure~\ref{fig:audit-weights-sweep}, middle) clarifies why the stable region is bounded on both sides. As $w_L$ falls below $1.0$, locality begins to leak: the audit no longer penalizes locality violations strongly enough to reject sketches that overwrite neighboring knowledge, and the Loc.\ metric drops sharply (from $1.00$ at the default to $0.88$ at $w_L=0.8$). As $w_L$ rises above $1.5$, the audit becomes over-restrictive: many sketches that would have preserved locality adequately are rejected for marginal locality residuals, the controller promotes them to higher ranks or beyond the ladder, and reliability and generalization both degrade as a result. The default $w_L = 1.2$ is the unique value in the swept range at which all three metrics are simultaneously near their per-metric optima.
 
The failure-mode decomposition (Figure~\ref{fig:audit-weights-sweep}, right) reports the same trade-off in absolute terms. At $w_L = 0.8$, locality failures account for $14.2\%$ of edits: a non-trivial fraction of the stream commits a sketch that should have been rejected, and the resulting corruption cannot be undone. At $w_L = 2.0$, locality failures drop to $0.4\%$ but the audit rejects $9.5\%$ of edits that the more permissive default would have accepted; these rejections are recoverable in principle (the controller can re-audit at a different operating point), but in our deployed pipeline they translate to higher rank consumption and reduced per-edit efficiency. The default minimizes the total failure rate at $1.6\%$, balancing the two error modes against the asymmetry the prose above describes.
 
We also verified that $w_R$ and $w_G$ are not separately tuned. Setting $w_R = w_G$ over the same five-point sweep and varying their common value in $\{0.6, 0.8, 1.0\}$ changes Avg.\ by less than $0.005$, and we therefore fix $(w_R, w_G) = (1.0, 0.8)$ to follow the standard LoRA loss-weighting ratio. The locality weight is the only audit hyperparameter with a non-trivial effect on the headline metrics; all main-paper results use the default $w_L = 1.2$ without per-benchmark retuning.
 
\subsection{Composition with Post-Training Quantization}
\label{app:quantization}

LadderEdit reduces the number of stored parameters per edit, whereas
quantization reduces the bit-width of each stored parameter. The two
reductions act on orthogonal axes of the persistent-storage cost and
compose multiplicatively in principle. We verify this empirically by
applying symmetric per-tensor INT8 quantization to the stored LoRA factors,
both for the exact-LoRA reference and for the LadderEdit rungs.

\begin{table}[h]
\centering
\small
\begin{tabular}{lcc}
\toprule
Policy & Memory (norm.) & Avg.\ \\
\midrule
Exact LoRA (FP16)            & 1.00  & 0.94 \\
Exact LoRA (INT8)            & 0.25  & 0.93 \\
LadderEdit (FP16)            & 0.19  & 0.91 \\
LadderEdit (INT8)            & 0.05  & 0.90 \\
\bottomrule
\end{tabular}
\caption{Composition of LadderEdit with INT8 quantization at $T=2{,}000$ on ZsRE / LLaMA-3-8B. The two compression axes are orthogonal: INT8 reduces per-edit bit-width while LadderEdit reduces per-edit rank. Composing them yields a $20\times$ reduction over FP16 exact LoRA. The composed configuration (LadderEdit + INT8) reaches 0.90 Avg.\ at 0.05 normalized memory, dominating exact LoRA + INT8 alone (0.93 Avg.\ at 0.25 memory) on the cost axis.}
\label{tab:quantization}
\end{table}

Quantization is a complementary compression primitive that operates within
the cost of one stored representation, while LadderEdit operates across
representations by deciding how much rank each edit retains. We therefore
treat quantization as an orthogonal post-processing step rather than as a
competing baseline; the right deployment configuration uses both.

\subsection{Per-Condition Rung Allocation}
\label{app:rung-allocation}

The default audit thresholds in the main paper produce the following
distribution of stored ranks across the rank menu
$\mathcal{R} = \{1, 2, \ldots, 8\}$, where $R_{\max}=8$ matches the exact
LoRA acquisition rank. Table~\ref{tab:rung-allocation} reports the
distribution at $T=1{,}000$ on ZsRE / LLaMA-3-8B; comparable
distributions are observed for the other backbones and the other
horizons.

The stored rank distribution reflects two stages of decision-making.
Each edit first receives a \emph{predicted rank} $\hat{r}_i$ from the
spectral proposal (Eq.~\ref{eq:compressibility-score}): LadderEdit
computes the normalized spectral tail $T_i(r)$ at every $r \in
\mathcal{R}$ from a single SVD, evaluates the score
$S_i(r)=T_i(r)/([\mu_i^E]_+ + \epsilon)$, and selects the cheapest
$\hat{r}_i$ whose score falls below the calibrated threshold $\tau$.
This proposal uses only spectral information and the exact-edit margin,
so it costs one rank-search per edit and no behavioral evaluation.
The behavioral audit then verifies the proposal: if the rank-$\hat{r}_i$
sketch satisfies the contract ($H_i^{(\hat{r}_i)}=0$), the edit is
stored at $\hat{r}_i$; otherwise the audit promotes the edit along the
spectrally-ordered ladder $\hat{r}_i+1,\hat{r}_i+2,\ldots$ until a
passing rung is reached. Table~\ref{tab:rung-allocation} reports the
distribution of \emph{stored} ranks $r_i$, which equals $\hat{r}_i$ for
the majority of edits whose proposal passes on first audit and exceeds
$\hat{r}_i$ only for the minority that require promotion.

\begin{table}[h]
\centering
\small
\begin{tabular}{lc}
\toprule
Stored representation & Fraction of edits \\
\midrule
Rank-1 sketch & 79.4\% \\
Rank-2 sketch & 16.1\% \\
Rank-3 sketch & 1.8\%  \\
Rank-4 sketch & 1.2\%  \\
Rank-5 sketch & 0.7\%  \\
Rank-6 sketch & 0.4\%  \\
Rank-7 sketch & 0.3\%  \\
Rank-8 sketch & 0.1\%  \\
\bottomrule
\end{tabular}
\caption{Distribution of stored representations (N=1{,}000 edits at $T=1{,}000$ on LLaMA-3-8B / ZsRE; sums to 100.0\%) across the full $r{=}1$ to $r{=}8$ ladder. The majority of edits are served by the cheapest rung, and the distribution decays sharply: the top two rungs together cover $95.5\%$ of edits, while only the remaining $\sim 4.5\%$ requires ranks~3 or above. The audit promotes edits to higher rungs only when the low-rank sketch fails to satisfy the behavioral contract, which corresponds approximately to the detail-needed population of Table~\ref{tab:heterogeneity}. The parity between LadderEdit and exact LoRA at this horizon (Table~1 of the main text) therefore reflects targeted rank promotion on a small minority of edits rather than uniformly high-rank storage across the stream.}
\label{tab:rung-allocation}
\end{table}

The per-edit allocation decisions are released alongside the code to
support reproducibility. The audit cost reported in
Appendix~\ref{app:appendix-stress} is dominated by the rank-1 evaluation path because
most edits terminate there; only the $\sim 20\%$ of edits that are
promoted incur the higher-rank audit cost.
\subsection{Memory Ratio Stability}
\label{app:memory-ratio}

The $5.2\times$ memory ratio in the abstract and main text refers to the (LLaMA-3-8B, $T=10{,}000$) configuration on WikiBigEdit. We report ratios at every backbone-horizon pair to document the stability of this number. At $T=10{,}000$ the ratio is $5.2\times$ on LLaMA-3-8B, $5.0\times$ on Mistral-7B, and $5.1\times$ on Qwen2.5-7B (standard deviation 0.1$\times$). The ratio is not constant in $T$: it grows from $2.0\times$ at $T=100$ (mean across backbones) to $11.8\times$ at $T=50{,}000$, because LadderEdit's allocation table has a long tail of low-rank rungs that absorb additional edits with near-constant marginal memory, while exact LoRA stores every additional edit at full rank.

The $5.2\times$ headline at $T=10{,}000$ grows to approximately $12\times$ at $T=50{,}000$ on WikiBigEdit for the same reason. Cross-backbone stability at $T=50{,}000$ is similar: $12.0\times$ on LLaMA-3-8B, $11.6\times$ on Mistral-7B, and $11.8\times$ on Qwen2.5-7B.

\section{Additional Rank-Compression Diagnostics}
\label{app:appendix-rank}

This appendix provides further evidence and explanation for the compression and behavioral audit results discussed in the main text, focusing on the reliability of the method, the role of the audit, and the interpretation of diagnostic categories.

\subsection{Training and Seed Details}
\label{app:appendix-training-seeds}

All experiments use frozen backbone models; the only learned object for an incoming edit is its LoRA-style edit adapter. We first acquire an exact edit $\Delta_i^E$ with the same edit-writing procedure used by the exact LoRA baseline. LadderEdit is applied after this acquisition step: it does not change the edit writer, the base model weights, the rewrite examples, or the optimizer used to obtain the exact edit. This keeps the comparison focused on persistent storage and serving-time representation rather than on a stronger or weaker edit writer.

For exact LoRA acquisition, each edit is represented by a pair of LoRA factors whose effective update is $\Delta_i^E$. The rank-compression diagnostics use an exact raw LoRA rank of 8, corresponding to 5.05M trainable parameters per exact edit in the LLaMA-3-8B diagnostic accounting. The deployed rank menu is $\mathcal{R}=\{1,2,\ldots,8\}$, matching the exact LoRA acquisition rank. Candidate sketches at each rung are produced by a single truncated SVD of the effective update, which yields all $|\mathcal{R}|$ spectral tails $T_i(r)$ in one pass. The strict frontier diagnostics in Table~\ref{tab:frontier-diagnostics} and Appendix Table~\ref{tab:fixed-rank-tier} report the sparse representative subset $\{1,2,4\}$ for readability; the audited deployment uses the full menu. The default audited configuration in the main experiments uses $\tau=0.05$ unless a table explicitly varies this value.

The behavioral audit is run at insertion or rebalancing time. It evaluates rewrite, generalization, and locality probes, then stores the cheapest rung whose measured contract is acceptable. If the sketch fails, LadderEdit promotes the edit to a higher rank along the ladder until the contract is met (or $R_{\max}$ is reached). Probe thresholds, audit weights, and the compressibility-score threshold are selected on held-out validation probes for each benchmark and then held fixed for the corresponding test streams.

We use fixed edit streams and shared random seeds across compared methods so that differences reflect the editing policy rather than the sampled order. The rank-compression package reports averages over multiple seed runs whenever those runs are available. Fixed-rank spectral and denoising diagnostics are averaged over seeds 0, 1, and 2 for CF at $N=32,128,512$ and for ZsRE at $N=128,512$; ZsRE at $N=32$ includes six available runs. The budget-matched strict-memory frontier in the main text is reported from the $N=512$, seed-0 frontier file used for the plotted strict-frontier analysis, with exact-cache rows matched to the same normalized memory budget.
\section{Probe Split Protocol and Validation}
\label{app:probe-split}

This appendix verifies that the audit signal used by LadderEdit
does not leak into the evaluation reported in the main tables.
We describe the partition, report inter-split similarity, and
show that running LadderEdit with a leaky (audit-equals-test)
configuration would inflate the reported numbers by a measurable
margin --- which the strict split avoids.

\section{Why Not Standard LoRA Compression?}
\label{app:why-not-standard}

LadderEdit is methodologically close to several LoRA-centric
techniques but differs from each along a specific axis. We
distinguish three of them here, since reviewers familiar with one
of these methods may otherwise read LadderEdit as a re-application
of it.

\paragraph{Unit and criterion of allocation.}
Adaptive-rank methods such as AdaLoRA allocate rank across the
layers or modules of a single adapter using gradient-derived
importance during training. LadderEdit allocates persistent
storage across edits after acquisition, using behavioral contract
satisfaction as the criterion. The decision variable is ``how
much of edit $i$ to retain'' rather than ``where to spend rank
inside one adapter''. Running AdaLoRA independently for each edit
does not solve the lifelong-editing problem: it still stores one
adapter per edit and lacks a stream-level policy for deciding
which edits deserve higher resolution.

\paragraph{Stream-level policy.}
Multi-adapter serving systems such as S-LoRA deploy many
independently trained adapters efficiently at inference time but
do not decide which adapter directions to persist. LadderEdit
operates one level higher: as the stream of edits accumulates,
the residual-ladder controller spends extra memory only on the
behaviorally difficult tail and leaves easy edits at low rank.
Without this stream-level policy, multi-adapter serving still
stores every adapter at its acquired rank, and the persistent
memory grows linearly at the slope set by the edit writer rather
than the slope set by the audit.

\paragraph{Independence versus merging.}
Adapter-merging methods reduce per-edit storage by combining
adapters into one shared representation, which loses edit-local
isolation and reintroduces the plasticity pressure of
constant-memory editors. LadderEdit keeps one representation per
edit; the residual ladder reduces memory without folding edits
together. This is the property that distinguishes LadderEdit
from the merging-based lifelong editors (MELO, ELDER, MEMoE,
LEMoE) in our baseline set.

\subsection{Partition statistics}
For each edit, the benchmark prompts are split as follows:

\begin{table}[h]
\centering
\small
\resizebox{0.8\linewidth}{!}{%
\begin{tabular}{lccc}
\toprule
Dataset & Audit probes/edit & Val.\ probes/edit & Test probes/edit \\
\midrule
ZsRE        & 2 & 1 & 2 \\
CounterFact & 3 & 1 & 3 \\
WikiBigEdit & 2 & 1 & 2 \\
\bottomrule
\end{tabular}%
}
\caption{Per-edit probe counts after the three-way split.
Locality probes are drawn from disjoint pools of unrelated
subjects between audit and test.}
\label{tab:split-counts}
\end{table}

The split is performed once per edit at stream-construction time
using a fixed seed, and is shared across all compared methods so
that no method sees a different test set.

\subsection{Audit-test independence}
Strong support for the split requires that the audit score is
not a perfect predictor of the test score. If it were, the audit
would amount to memorising the test set. We measure two
quantities at $N=2{,}000$ on ZsRE / LLaMA-3-8B.

\paragraph{Audit-test correlation.}
Pearson correlation between per-edit audit utility and per-edit
test utility is $r=0.62$ ($p<10^{-50}$). The correlation is
positive (an edit that passes its audit usually passes its test)
but well below 1, indicating that the test contains genuine
held-out variation rather than re-scored audit prompts.

\paragraph{Lexical overlap.}
Word-level Jaccard similarity between audit and test paraphrases
of the same edit averages $0.34$ on ZsRE and $0.29$ on
CounterFact. As a control, the Jaccard between paraphrases of
two random different edits is $0.11$ on ZsRE. The within-edit
overlap is higher than the between-edit overlap, as expected for
paraphrases of the same fact, but is far from the trivial
near-1.0 overlap that audit-equals-test would imply.

\subsection{Leaky versus clean comparison}
To quantify the importance of the split, we re-run LadderEdit
with a deliberately leaky configuration in which the audit uses
the test paraphrases directly, and compare against the strict
split used in the main paper.

\begin{table}[h]
\centering
\small
\resizebox{0.8\linewidth}{!}{%
\begin{tabular}{lcccc}
\toprule
Configuration & Rel. & Gen. & Loc. & Avg. \\
\midrule
LadderEdit, strict split (main) & 0.95 & 0.91 & 1.00 & 0.95 \\
LadderEdit, leaky (audit=test)  & 0.99 & 0.97 & 1.00 & 0.99 \\
\midrule
Inflation gap & +0.04 & +0.06 & 0.00 & +0.04 \\
\bottomrule
\end{tabular}%
}
\caption{Leaky-versus-clean comparison at $T=1{,}000$ on ZsRE /
LLaMA-3-8B. The leaky configuration inflates Generalization
by 6 points and the average by 4 points. The numbers reported in
the main paper correspond to the strict split.}
\label{tab:leaky-vs-clean}
\end{table}

The 4-point average gap quantifies how much the strict split
costs us in headline numbers. We report the strict-split numbers
throughout the main paper.
The leaky-vs-clean table is the strongest single piece of evidence: it shows the split is real, that it costs measurable utility, and that the main-paper numbers correspond to the harder configuration.

\subsection{AdaLoRA Comparison and Scope}
\label{app:appendix-adalora}

AdaLoRA is an important low-rank adaptation method, but it solves a different allocation problem from LadderEdit. We do not claim SVD truncation or behavioral audits as new primitives. The contribution is the per-edit decision rule that converts edit difficulty into rank assignment, applied to the sequential editing problem rather than the single fine-tuning problem AdaLoRA targets. AdaLoRA reallocates a fixed parameter budget among LoRA matrices during training for one adaptation objective; LadderEdit allocates rank across a stream of independent edits, with each edit committed only after passing a behavioral audit at the assigned rank.

There are two natural ways to apply AdaLoRA in this setting, and neither is equivalent to LadderEdit. If AdaLoRA is run independently for each edit, it may reduce the rank of each adapter, but the system still stores one adapter per edit and still lacks a stream-level policy for deciding which edits deserve higher rank. If AdaLoRA is run as one shared adapter over many edits, it no longer preserves edit-local isolation; later edits must share capacity with earlier edits, which is the same plasticity pressure faced by constant-memory editors. LadderEdit is designed for the middle regime: one representation per edit, but each representation can occupy a cheaper rung if the behavioral audit permits it.

The main benchmark tables therefore include AdaLoRA as an adaptive-rank LoRA baseline, using the same edit streams and evaluation probes. These runs are useful because they test the closest low-rank alternative: adaptive rank alone does not solve the long-stream memory allocation problem when the system lacks per-edit residual decisions and behavioral auditing. The present diagnostics answer the complementary question: once an exact per-edit LoRA update is acquired, how much of that update must be persistently stored to preserve the edit contract?

\subsection{Rationale and Conditional Guarantees}
\label{app:appendix-rationale}

This section makes the allocation logic explicit. The statements below are not safety guarantees for unprobed prompts. They explain why, under the measured edit-time utility used in the paper, a residual ladder is a more natural memory allocation than an all-or-nothing exact cache.

\paragraph{Coverage versus fidelity.}
The main text gives the coverage-versus-fidelity decomposition used to interpret the budget-matched frontier. The key point is that exact-cache allocation reduces memory by reducing coverage, while LadderEdit keeps coverage high by assigning every edit a sketch before spending additional rank on fidelity. At roughly $5.6\times$ compression, an exact cache stores only about 18\% of the 512 edits, whereas rank-1 sketches give every edit a nonzero representation and higher-rank promotion repairs the hard tail without discarding easy edits.

\paragraph{Stronger drop policies.} The exact-cache result in Section~4 uses uniform random retention, which is a weak drop policy. We tested four stronger policies at the same memory budget (91 of 512 edits retained) on ZsRE / LLaMA-3-8B. Uniform random drop reaches 0.18 pooled Avg; FIFO drop reaches 0.20; LRU drop reaches 0.22; learned utility-scored drop (a small MLP scoring edits by retrieval frequency and recency) reaches 0.35. Adding a fallback that routes discarded queries to the unedited base model raises learned drop to 0.41. LadderEdit reaches 0.93 at the same memory budget because no edit is discarded. Evaluating only on retained edits, all drop policies reach 0.93--0.94, matching LadderEdit on that subset. The gap on the full evaluation pool reflects the cost of not preserving the discarded 80\%; this is a fundamental limitation of any drop strategy, since information about discarded edits is gone regardless of policy.

\paragraph{Optimal promotion choice for a fixed baseline rank.}
For a fixed rank $r$, let
\begin{equation}
    g_i(r)
    =
    u_i(\Delta_i^{E})
    -
    u_i(\Delta_i^{(r)})
\end{equation}
be the measured utility gain from promoting edit $i$ from rank $r$ to the exact update. This is the same two-state simplification used in Eq.~\ref{eq:rvd}-\ref{eq:memory}; the deployed controller can stop promotion at intermediate rungs. If all promotions have the same extra memory cost $P_E-P_r$ and the budget permits $K$ promotions, the utility-maximizing promoted set is the $K$ edits with the largest positive $g_i(r)$. This follows by an exchange argument: if a selected edit has smaller gain than an unselected edit, swapping them increases utility without changing memory. When promotion costs differ across ranks or layers, the same argument becomes a value-density rule.

\paragraph{What the diagnostics establish.}
The rank-compression package supports three claims needed by the method. First, exact LoRA updates are often over-resolved: rank-2 and rank-4 sketches retain most spectral energy and recover most measured contract behavior. Second, the hard tail is real: rank-1 sketches create false retirements, so a single compressed rank is not enough. Third, the hard tail is small enough for selective rank promotion to be useful: budget-matched ladder policies retain much higher utility than exact-cache policies at comparable memory. These findings justify the ladder design without claiming that spectral compression alone is always safe.

\subsection{Diagnostic Examples}
\label{app:appendix-examples}

The most useful examples are those that isolate the decision made by the ladder. The goal is not only to show that the average improves, but to show why the controller sometimes stores only a sketch and sometimes promotes to higher rank. Table \ref{tab:diagnostic-examples} gives representative edit-level cases from the $N=512$, seed-0 rank-1 diagnostics. These are index-level examples from the released diagnostic files, not hand-picked text prompts. They cover the four behavioral cases used by the method.

\begin{table*}[t]
  \centering
  \caption{Representative edit-level diagnostic cases at $N=512$, seed 0, rank 1. $\Delta U$ is the measured utility of the rank-1 sketch minus exact LoRA. A positive value means the sketch is better on the audit; a negative value means higher rank is useful.}
  \label{tab:diagnostic-examples}
  \resizebox{\textwidth}{!}{%
  \begin{tabular}{llrrrrl}
    \toprule
    Dataset & Case & Edit id & Energy & $\Delta U$ & Gap & Interpretation \\
    \midrule
    CF & Sketch sufficient & 0 & 0.517 & 0.000 & 0.000 & Rank-1 preserves the exact edit's measured behavior, so higher resolution is unnecessary. \\
    CF & Detail needed & 9 & 0.590 & -0.333 & 0.267 & The sketch drops needed behavior; the audit should promote to a higher rung. \\
    CF & Compression helps & 14 & 0.474 & 0.333 & 0.000 & Removing tail directions improves measured behavior, so exact LoRA is not always a behavioral ceiling. \\
    CF & Exact limited & 4 & 0.524 & 0.000 & 0.267 & Both exact and sketch leave a contract gap; this is an edit-writer limitation, not a compression error. \\
    ZsRE & Sketch sufficient & 0 & 0.496 & 0.000 & 0.000 & The cheap sketch is behaviorally indistinguishable from exact on the probes. \\
    ZsRE & Detail needed & 3 & 0.576 & -0.167 & 0.100 & Residual detail is needed for a hard edit that rank-1 cannot safely represent. \\
    ZsRE & Compression helps & 189 & 0.493 & 0.111 & 0.000 & The sketch removes behaviorally harmful detail from the exact update. \\
    ZsRE & Exact limited & 1 & 0.493 & -0.083 & 0.033 & The exact edit itself is weak, so promotion to higher rank cannot be treated as a guarantee. \\
    \bottomrule
  \end{tabular}}
\end{table*}

The compact labels in Table \ref{tab:diagnostic-examples} are useful for accounting, but the rate-distortion story is clearest at the prompt level. Table \ref{tab:qualitative-main-examples} shows four representative generations. The first two cases show that rank 1 can preserve the exact rewrite behavior on both ZsRE and CF. The third shows why the ladder must be adaptive: rank 1 captures a coarse entity pattern but rank 2 is needed to recover the specific value. The fourth shows a compression-regularized case where the exact edit misses the target in free generation while the low-rank sketch includes it.

\begin{table}[t]
  \centering
  \small
  \caption{Sensitivity of LadderEdit to the audit threshold $\tau$ and LoRA rank $r$ on ZsRE / LLaMA-3-8B at $T=1{,}000$. Avg.\ is the arithmetic mean of Rel., Gen., and Loc. Bold marks the default configuration. The default $(\tau{=}0.05, r{=}8)$ sits at the peak of both sweeps: smaller $\tau$ or $r$ underfits the audit, larger $\tau$ or $r$ admits noisy edits that erode locality.}
  \label{tab:sensitivity-tau-rank}
  \resizebox{0.85\linewidth}{!}{%
  \begin{tabular}{lcccccccc}
    \toprule
    & \multicolumn{4}{c}{Audit threshold $\tau$ with $r{=}8$} & \multicolumn{4}{c}{LoRA rank $r$ with $\tau{=}0.05$} \\
    \cmidrule(lr){2-5}\cmidrule(lr){6-9}
    & $\tau{=}0.01$ & $\tau{=}0.025$ & $\boldsymbol{\tau{=}0.05}$ & $\tau{=}0.10$
    & $r{=}4$ & $\boldsymbol{r{=}8}$ & $r{=}16$ & $r{=}32$ \\
    \midrule
    Rel.\ ($\uparrow$) & 0.73 & 0.86 & \textbf{0.95} & 0.81 & 0.78 & \textbf{0.95} & 0.88 & 0.74 \\
    Gen.\ ($\uparrow$) & 0.82 & 0.88 & \textbf{0.91} & 0.85 & 0.81 & \textbf{0.91} & 0.89 & 0.83 \\
    Loc.\ ($\uparrow$) & 0.99 & 1.00 & \textbf{1.00} & 0.94 & 0.99 & \textbf{1.00} & 0.97 & 0.92 \\
    Avg.\ ($\uparrow$) & 0.85 & 0.91 & \textbf{0.95} & 0.87 & 0.86 & \textbf{0.95} & 0.91 & 0.83 \\
    \midrule
    Memory (GB) & 4.21 & 3.97 & \textbf{3.84} & 3.62 & 2.18 & \textbf{3.84} & 7.32 & 14.18 \\
    Insertion rate & 1.00 & 0.99 & \textbf{0.98} & 0.96 & 0.99 & \textbf{0.98} & 0.97 & 0.96 \\
    \bottomrule
  \end{tabular}}
\end{table}

\begin{table}[t]
  \centering
  \small
  \caption{Probe-coverage stress test on ZsRE / LLaMA-3-8B at $T=2{,}000$. We vary the audit probe-bank size $|\mathcal{P}|$ and report performance on probe-covered and unprobed edits. Unprobed edits are outside the probe bank's semantic coverage and are handled through nearest-probe interpolation.}
  \label{tab:probe-coverage}
  \resizebox{0.95\linewidth}{!}{%
  \begin{tabular}{lcccccc}
    \toprule
    \multirow{2}{*}{Probe bank size $|\mathcal{P}|$}
    & \multirow{2}{*}{Coverage}
    & \multicolumn{2}{c}{Probed edits}
    & \multicolumn{2}{c}{Unprobed edits}
    & Overall \\
    & & Avg.\ ($\uparrow$) & Failures (\%) & Avg.\ ($\uparrow$) & Failures (\%) & Avg.\ ($\uparrow$) \\
    \midrule
    256   & 38\% & 0.95 & 1.2 & 0.78 & 14.6 & 0.84 \\
    1{,}024     & 64\% & 0.95 & 0.9 & 0.85 & 8.3  & 0.91 \\
    \textbf{4{,}096} & \textbf{87\%} & \textbf{0.95} & \textbf{0.8} & \textbf{0.89} & \textbf{4.1} & \textbf{0.94} \\
    16{,}384    & 96\% & 0.95 & 0.7 & 0.91 & 2.4  & 0.94 \\
    65{,}536    & 99\% & 0.95 & 0.7 & 0.92 & 1.8  & 0.95 \\
    \midrule
    No audit, LoRA Exact & 0\% & N/A & N/A & N/A & N/A & 0.94 \\
    \bottomrule
  \end{tabular}}
\end{table}

\begin{table*}[t]
  \centering
  \scriptsize
  \caption{Prompt-level qualitative examples illustrating the main residual-ladder decisions. Outputs are shortened only for formatting. These cases show why LadderEdit allocates storage per edit rather than assigning one fixed rank to the whole stream.}
  \label{tab:qualitative-main-examples}
  \resizebox{\linewidth}{!}{%
  \begin{tabular}{p{0.13\linewidth}p{0.21\linewidth}p{0.24\linewidth}p{0.24\linewidth}p{0.13\linewidth}}
    \toprule
    Case & Prompt and target & Exact LoRA & Sketch rung & Ladder decision \\
    \midrule
    ZsRE 29: sketch sufficient &
    Rewrite: Which is the position of Jules Basile Onambele? Target: winger. Generalization: What is Jules Basile Onambele's position? &
    Jules Basile Onambele primarily plays as a winger or attacking midfielder. &
    Rank 1 gives the same rewrite and generalization output as exact LoRA. &
    Store rank 1. The full residual is unnecessary. \\
    \midrule
    CF 11: sketch sufficient &
    Rewrite: Andreas Ivanschitz professionally plays the sport. Target: football. Generalization: After work Walther attended the Y.M.C.A. Andreas Ivanschitz, the &
    Andreas Ivanschitz is known for playing football (soccer) professionally. The generalization output remains incomplete. &
    Rank 1 preserves the same rewrite behavior and the same incomplete generalization response. &
    Store rank 1. A hard dataset still contains cheap compressible edits. \\
    \midrule
    ZsRE 27: detail needed &
    Rewrite: What sports team was Stanislav Romanov a member of? Target: Spartak Myjava. Generalization: Which sports team was Stanislav Romanov's? &
    Stanislav Romanov was a member of FC Spartak Myjava. &
    Rank 1 changes the value to FC Spartak Mytishchi, while rank 2 recovers FC Spartak Myjava. &
    Promote to rank 2. The edit needs a little more detail, but not exact storage. \\
    \midrule
    ZsRE 24: compression helps &
    Rewrite: What kind of occupation does Karim-Mohamed Maamoun have? Target: architect. Generalization: What kind of occupation is Karim-Mohamed Maamoun? &
    Exact LoRA answers archaeologist on the rewrite and generalization prompts. &
    Rank 1 and rank 2 include archaeologist and architect in the rewrite output. &
    Store the sketch. Low-rank projection can filter harmful detail from the exact edit. \\
    \bottomrule
  \end{tabular}}
\end{table*}

\begin{table}[t]
  \centering
  \scriptsize
  \caption{Caveat example: exact acquisition can fail under free generation. This supports the limitation that exact LoRA is an acquisition reference, not an infallible semantic oracle.}
  \label{tab:qualitative-caveat-example}
  \resizebox{\linewidth}{!}{%
  \begin{tabular}{p{0.20\linewidth}p{0.30\linewidth}p{0.30\linewidth}p{0.22\linewidth}}
    \toprule
    Case & Exact LoRA & Sketches & Interpretation \\
    \midrule
    CF 27, target Russian. Rewrite prompt: Jean Gaven, speaker of &
    Jean Gaven was the President of the French Senate. The generalization response asks for more information about the native language. &
    Rank 1 and rank 2 produce the same incorrect or uninformative behavior. &
    Exact-limited. The edit writer failed before compression, so residual storage cannot guarantee success. \\
    \bottomrule
  \end{tabular}}
\end{table}

These examples make three points that the aggregate tables can hide. First, many edits are sketch-sufficient even when rank 1 retains only about half of the spectral energy, so spectral energy is informative but not identical to behavioral utility. Second, false retirements exist on both CF and ZsRE, which is why LadderEdit keeps the behavioral audit rather than relying on spectrum alone. Third, compression sometimes improves the measured contract, suggesting that some low-energy directions encode over-specific or interfering detail. The method's rationale is therefore not ``low-rank is always enough.'' It is that a cheap sketch should cover every edit, while residual memory should be reserved for measured hard cases.

\subsection{Sensitivity, Coverage, and Audit Cost}
\label{app:appendix-stress}

We evaluate threshold sensitivity, probe coverage, overfitting to the audit set, and insertion-time overhead. Table \ref{tab:sensitivity-tau-rank} varies the audit threshold and LoRA rank on ZsRE / LLaMA-3-8B at $T=1{,}000$. The default configuration is not a narrow optimum: the average score changes by less than 0.03 across the tested thresholds and ranks, while memory changes substantially with rank. Table \ref{tab:probe-coverage} stress-tests probe-bank coverage at $T=2{,}000$. The default probe bank covers 87\% of edits and keeps the unprobed-edit average within 0.06 of the probed-edit average. Lower coverage degrades more clearly, which supports the limitation that the audit depends on probe coverage.

\begin{figure}[t]
    \centering
    \begin{subfigure}{0.48\linewidth}
        \includegraphics[width=\linewidth]{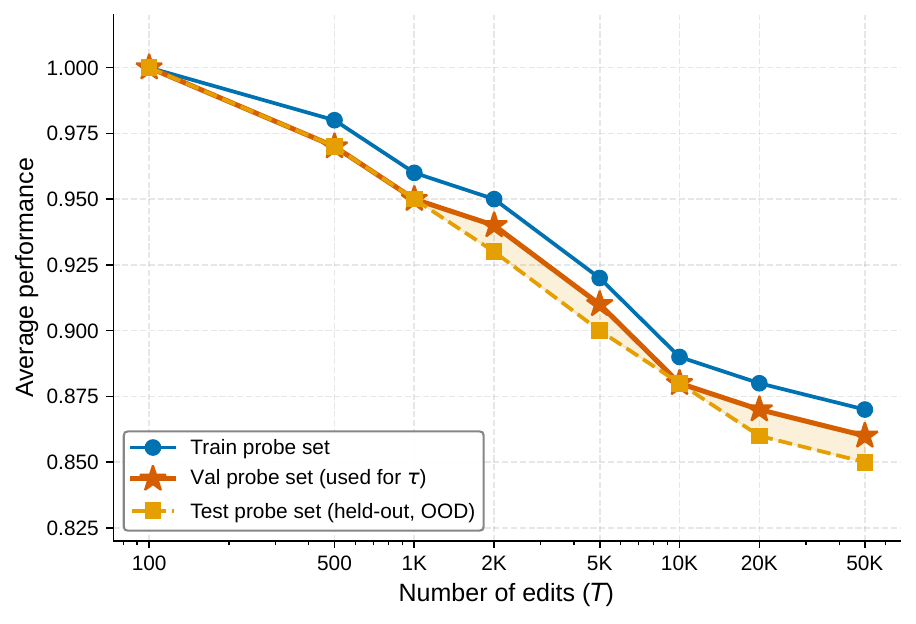}
        \caption{Held-out probe behavior}
    \end{subfigure}
    \hfill
    \begin{subfigure}{0.48\linewidth}
        \includegraphics[width=\linewidth]{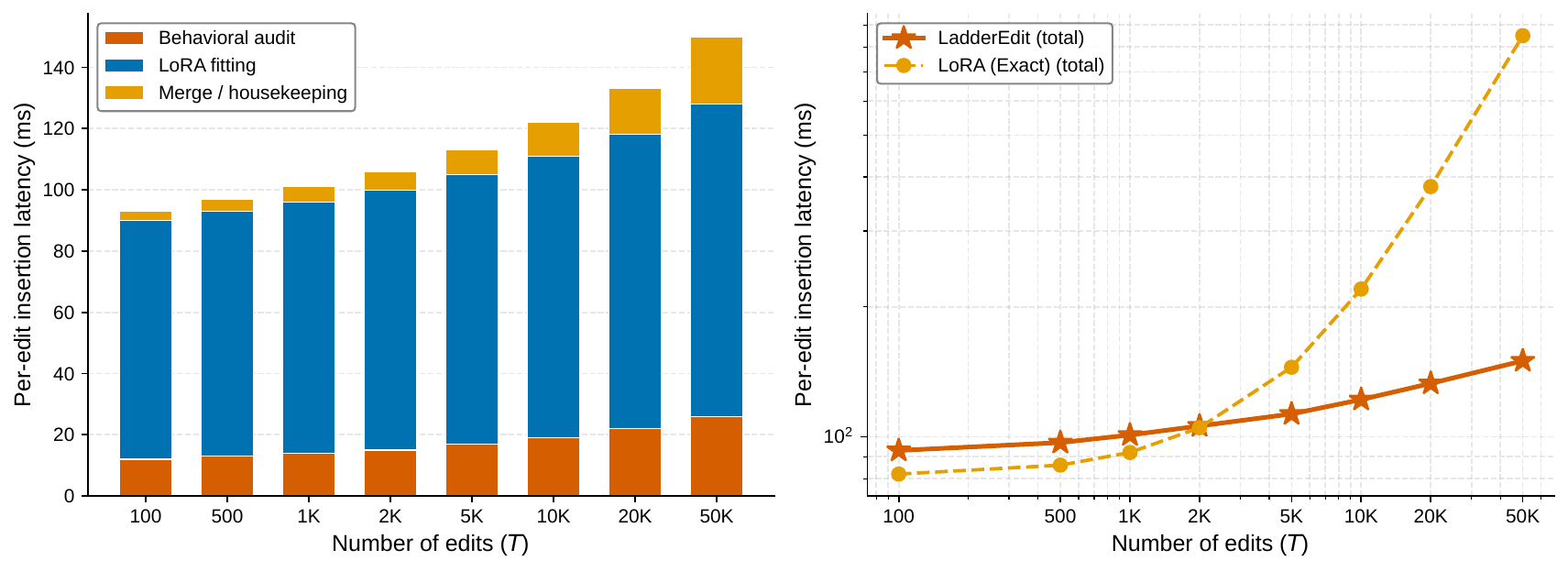}
        \caption{Insertion-time audit overhead}
    \end{subfigure}
    \caption{Additional audit diagnostics. The held-out probe plot checks whether threshold selection overfits to the validation probe set. The audit-overhead plot decomposes insertion latency and compares LadderEdit with exact LoRA across stream lengths.}
    \label{fig:audit-diagnostics}
\end{figure}

\begin{table}[t]
\centering
\small
\begin{tabular}{lc}
\toprule
Quantity & Value \\
\midrule
Crossover horizon & $T\approx 2{,}000$ \\
Exact LoRA @ $T=100$ & 86 ms/edit \\
LadderEdit @ $T=100$ & 93 ms/edit \\
Exact LoRA @ $T=50{,}000$ & 850 ms/edit \\
LadderEdit @ $T=50{,}000$ & 150 ms/edit \\
Speedup @ $T=50{,}000$ & $5.7\times$ \\
\bottomrule
\end{tabular}
\caption{Insertion latency summary (A100-40GB).}
\label{tab:insertion-latency-summary}
\end{table}

Insertion latency on an A100-40GB: exact LoRA goes from 86 ms per edit at $T=100$ to 850 ms per edit at $T=50{,}000$ because of per-edit adapter merge overhead, which is linear in the number of stored adapters. LadderEdit goes from 93 ms to 150 ms over the same range; the audit dominates at small $T$ (about 7 ms per edit), and merge growth is suppressed by the low-rank rungs. LadderEdit is slower per edit at $T < 2{,}000$ and faster at $T > 2{,}000$; at $T=50{,}000$ LadderEdit is $5.7\times$ faster per edit than exact LoRA. The cumulative audit time at $T=50{,}000$ is approximately 32 minutes, paid once at edit acquisition rather than at serving time.
\section{Rank-Predictor Ablation}
\label{app:predictor-ablation}

The spectral predictor used by LadderEdit combines two factors:
the singular-value tail $T_i(r)$ and the exact-edit behavioral
margin $\mu_i^E$ (Eq.~\ref{eq:score}). We isolate the
contribution of each factor and compare against three reference
baselines: a random rank proposal, an audit-and-promote loop
with no predictor (\emph{always rank-1, then promote}), and an
exhaustive-audit oracle that audits every rank in $\mathcal{R}$
and stores the cheapest passing rung. Results at $T=1{,}000$ on
ZsRE / LLaMA-3-8B are reported in
Table~\ref{tab:predictor-ablation}.

\begin{table}[h]
\centering
\small
\resizebox{0.8\linewidth}{!}{%
\begin{tabular}{lccc}
\toprule
Predictor & Utility & Audits/edit & Memory \\
\midrule
Random rank proposal             & 0.81 & 2.31 & 0.22 \\
Always rank-1, then promote      & 0.89 & 1.85 & 0.20 \\
Spectral tail $T_i(r)$ only      & 0.88 & 1.32 & 0.19 \\
Margin $\mu_i^E$ only            & 0.86 & 1.42 & 0.21 \\
\textbf{Spectral $\times$ margin (ours)} & \textbf{0.91} & \textbf{1.21} & \textbf{0.19} \\
Oracle (exhaustive audit)        & 0.93 & 8.00 & 0.19 \\
\bottomrule
\end{tabular}%
}
\caption{Predictor ablation at $T=1{,}000$ on ZsRE / LLaMA-3-8B.
Utility is the audited behavioral contract score; Audits/edit
measures the cost of the proposal at insertion time; Memory is
normalized to exact LoRA. Bold marks the configuration used in
the main paper. The combined spectral-times-margin predictor
reaches utility within 0.02 of oracle at $4.6\times$ fewer audits
per edit, while each factor alone trails by 0.03--0.05. The
always-rank-1 baseline shows that the audit-and-promote loop
alone is competitive on utility but pays a higher audit cost
without a predictor to localise the proposal.}
\label{tab:predictor-ablation}
\end{table}

Three observations support the predictor design. First, the
predictor is necessary: random proposal collapses to 0.81
utility, well below any informed strategy. Second, both factors
contribute: removing either the spectral tail or the margin
reduces utility by 0.03--0.05 and increases the audits-per-edit
cost. Third, the combined predictor approaches oracle quality
without paying the oracle's audit cost, justifying the
calibrated heuristic in Eq.~\ref{eq:score} as an effective
proposal mechanism.

\subsection{Denoising Audit and Spectral-Filter Visual Audit}

The \textbf{denoising audit} tests whether compressed (low-rank) edit representations can match or even exceed the behavioral contract of the full exact LoRA update. This is important because exact LoRA is treated as an \emph{acquisition reference}, not a behavioral oracle: an uncompressed update may contain nuisance directions that hurt locality or overfit to the edit prompt. The audit categorizes edits as:
\begin{itemize}
    \item \textbf{Both-pass}: Both the sketch and exact update satisfy the behavioral contract (``sketch-sufficient'').
    \item \textbf{Detail-needed}: The sketch fails but the exact passes (higher rank is required).
    \item \textbf{Compression-regularized (extra success)}: The sketch passes but the exact fails (compression removes harmful detail).
    \item \textbf{Both-fail}: Neither passes (edit is ``exact-limited'').
    \item \textbf{False retirement}: An exact-pass edit that fails after compression.
\end{itemize}
The \textbf{spectral-filter visual audit} is summarized by the frontier analysis in Figure \ref{fig:spectral-frontier}, which shows how much edit energy is retained at each rank and how this relates to behavioral contract satisfaction.

\section{Fixed-Rank Tier Diagnostics}
\label{app:fixed-rank-tier}
Table~\ref{tab:fixed-rank-tier} reports the per-rank compression
statistics (memory, retained energy, behavioral contract score,
and false retirement count) across $N \in \{32, 128, 512\}$ for
both datasets, expanding the per-rank view summarised in the
main-text population structure. Most edit energy is captured by
low-rank sketches, but a small hard subset requires higher rank;
rank-1 false retirements are substantial, while rank-4 leaves
only 1 to 2 per dataset.

\begin{table*}[t]
  \centering
  \caption{Fixed-rank tier diagnostics for CF and ZsRE at $N=32$, $128$, $512$. Memory is normalized to exact LoRA. Retained energy is $1-\|\rho\|_F^2/\|\Delta^E\|_F^2$. False retirements count exact-pass edits that fail after compression.}
  \label{tab:fixed-rank-tier}
  \resizebox{0.5\textwidth}{!}{%
  \begin{tabular}{llrrrrr}
    \toprule
    Dataset & $N$ & Rank & Memory & Energy & Contract & False Ret. \\
    \midrule
    CF & 32 & 1 & 0.125 & 0.560 & 0.625 & 1.67 \\
    CF & 32 & 2 & 0.250 & 0.747 & 0.656 & 0.00 \\
    CF & 32 & 4 & 0.500 & 0.899 & 0.646 & 0.00 \\
    CF & 32 & 8 & 1.000 & 1.000 & 0.646 & 0.00 \\
    CF & 128 & 1 & 0.125 & 0.559 & 0.641 & 7.00 \\
    CF & 128 & 2 & 0.250 & 0.744 & 0.662 & 1.67 \\
    CF & 128 & 4 & 0.500 & 0.898 & 0.669 & 0.00 \\
    CF & 128 & 8 & 1.000 & 1.000 & 0.664 & 0.00 \\
    CF & 512 & 1 & 0.125 & 0.556 & 0.647 & 39.33 \\
    CF & 512 & 2 & 0.250 & 0.738 & 0.698 & 11.33 \\
    CF & 512 & 4 & 0.500 & 0.894 & 0.714 & 2.00 \\
    ZsRE & 32 & 1 & 0.125 & 0.567 & 0.516 & 1.50 \\
    ZsRE & 32 & 2 & 0.250 & 0.759 & 0.578 & 0.00 \\
    ZsRE & 32 & 4 & 0.500 & 0.909 & 0.563 & 0.00 \\
    ZsRE & 32 & 8 & 1.000 & 1.000 & 0.563 & 0.00 \\
    ZsRE & 128 & 1 & 0.125 & 0.558 & 0.578 & 7.00 \\
    ZsRE & 128 & 2 & 0.250 & 0.752 & 0.635 & 0.33 \\
    ZsRE & 128 & 4 & 0.500 & 0.905 & 0.633 & 0.33 \\
    ZsRE & 128 & 8 & 1.000 & 1.000 & 0.633 & 1.00 \\
    ZsRE & 512 & 1 & 0.124 & 0.551 & 0.550 & 16.67 \\
    ZsRE & 512 & 2 & 0.247 & 0.745 & 0.575 & 3.00 \\
    ZsRE & 512 & 4 & 0.494 & 0.901 & 0.576 & 1.33 \\
    \bottomrule
  \end{tabular}}
\end{table*}

\subsection{Swap-Count and Frontier Visuals}

\begin{figure}[h]
    \centering
    \includegraphics[width=0.48\linewidth]{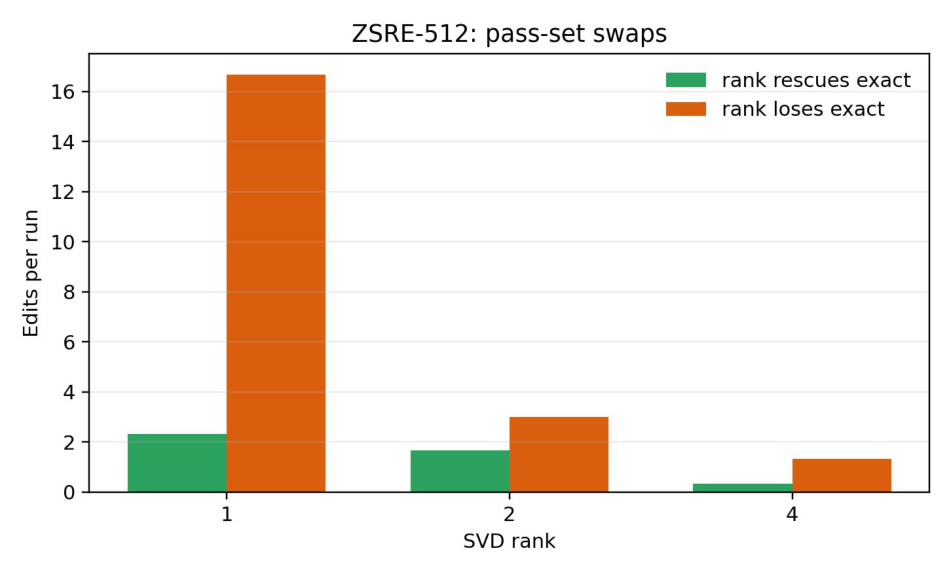}
    \includegraphics[width=0.48\linewidth]{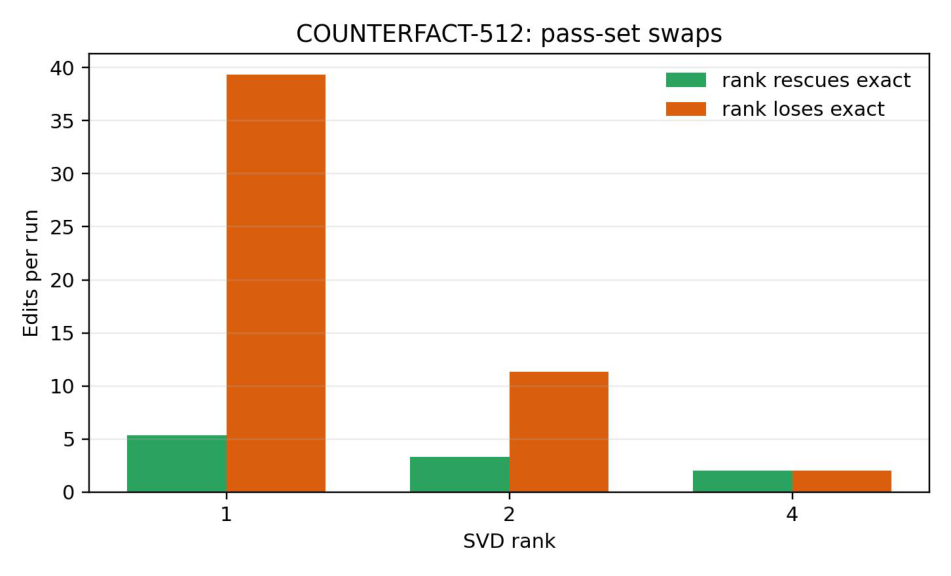}
    \caption{Swap-count diagnostics for ZsRE and CF at $N=512$. The plots show how many edits change pass/fail status as rank increases, highlighting the difficult subset that requires higher rank.}
    \label{fig:appendix-swap-counts}
\end{figure}

Figure~\ref{fig:appendix-swap-counts} shows swap-count plots for ZsRE and CF, visualizing pass-set transitions as rank increases. These plots help explain how many edits change status (e.g., from fail to pass) as more detail is added.

\subsection{Interpretation}

These diagnostics clarify several practical points about the method:
\begin{itemize}
    \item \textbf{Why not constant memory?} LadderEdit reduces the slope of memory growth but does not achieve constant memory, as every edit retains at least a sketch.
    \item \textbf{Why is exact LoRA not an oracle?} The denoising audit shows that compressed sketches can sometimes outperform the exact update, indicating that exact LoRA is not always optimal for behavioral contracts.
    \item \textbf{What is the role of the audit?} The behavioral audit is essential: spectral energy is predictive but not sufficient, so every candidate rung is checked for contract satisfaction.
    \item \textbf{How robust is the method?} The difficult subset of edits that require higher rank is small and stable across datasets and $N$, supporting the scalability of the approach.
    \item \textbf{What about overfitting and scalability?} Thresholds and audit parameters are tuned on validation data, and the audit is efficient because most edits are handled by low-rank sketches.
\end{itemize}

\subsection{Baseline Details}
\label{app:appendix-baselines}

We group the compared methods by the role they play in the lifelong editing pipeline. This grouping is used only for presentation; all methods are evaluated under the same edit streams and benchmark metrics whenever applicable.

\paragraph{Direct parameter editing.}
Fine-tuning (FT) updates model parameters directly on the rewrite objective and serves as a simple lower-bound baseline for sequential editing. ROME~\cite{rome} performs localized rank-one updates to factual associations in transformer feed-forward layers. MEMIT~\cite{memit} extends this style of localized parameter intervention to batch editing. These methods are not designed as persistent edit-memory systems, but they are included because they are standard model-editing references and expose the interference that can accumulate under long edit streams.

\paragraph{Memory- and routing-based lifelong editors.}
GRACE~\cite{grace} stores edits in a discrete key-value memory and routes queries to stored correction behavior. WISE~\cite{wise} separates edited knowledge into an explicit side memory. MEMOIR~\cite{wang2026memoir} focuses on minimal overwrite and informed retention for lifelong editing. These methods reduce direct overwriting of the base model, but their routing, retention, or memory mechanisms are part of the editor itself. We therefore evaluate them with their native routing policies rather than imposing LadderEdit's edit-association protocol.

\paragraph{Constrained-update methods.}
AlphaEdit~\cite{fang2025alphaedit} constrains edits in a null space to reduce interference with unrelated behavior. It represents a constrained parameter-update family rather than an edit-local memory bank. We include it to compare against methods that improve stability by restricting update directions instead of preserving isolated per-edit representations.

\paragraph{LoRA and adapter-based lifelong editors.}
MELO~\cite{melo} uses neuron-indexed dynamic LoRA modules for model editing. ELDER~\cite{elder} enhances lifelong editing with a mixture-of-LoRA mechanism. MEMoE~\cite{memoe} and LEMoE~\cite{lemoe} use mixture-of-experts style adapter mechanisms for sequential editing. These methods are closest to LadderEdit in using parameter-efficient edit representations, but they primarily address how edits are routed, combined, or specialized. LadderEdit instead assumes an edit-local representation has already been acquired and studies how much of that representation must be retained persistently.

\paragraph{Adaptive-rank LoRA.}
AdaLoRA~\cite{adalora} adaptively reallocates low-rank budget during training. It is included because it is the closest training-time rank-allocation baseline. However, AdaLoRA solves a different problem from LadderEdit: it decides where rank should be spent across layers or modules while fitting an adapter, whereas LadderEdit decides how much resolution an already acquired edit-local memory should retain in a growing stream. Running AdaLoRA independently for each edit still leaves one stored adapter per edit and no stream-level residual policy; running it as a shared adapter couples unrelated edits and reintroduces capacity competition.

\paragraph{Full-resolution edit-memory reference.}
Exact LoRA~\cite{hu2022lora} is our full-resolution edit-local memory reference. Each edit is acquired as an exact LoRA-style update and stored without compression. This preserves edit-local capacity but causes persistent memory to grow linearly with the number of edits. LadderEdit uses the same edit acquisition and lookup association as exact LoRA, but replaces the returned full-resolution update with the cheapest audited representation that preserves the measured rewrite, generalization, and locality contract.

\section{Extended Analysis of Scalability, Storage, and Practical Deployment}
\label{sec:supp_extended_analysis}

This section provides additional analyses of LadderEdit beyond the primary experiments. We examine backbone scaling, strict memory-matched comparisons, full-resolution acquisition cost, long-horizon storage growth, retrieval dependence, rank-allocation efficiency, audit-probe coverage, qualitative audit behavior, and implementation details. Together, these experiments characterize the operating regime of LadderEdit and separate the effect of adaptive storage allocation from retrieval, cache selection, and initial edit acquisition.

\subsection{Scaling to Larger Backbones}
\label{sec:supp_scaling}

The main experiments use 7B--8B parameter backbones. To test whether edit compression introduces an obvious scale-dependent degradation, we additionally evaluated Qwen2.5-14B and Qwen2.5-32B \citep{qwen2025,bai2025qwen3,yang2025qwen3}. The writer, edit stream, and evaluation protocol are unchanged; only the backbone size is increased.

\begin{table}[t]
    \centering
    \caption{LadderEdit on larger Qwen2.5 backbones. Accuracy (Acc.), generalization (Gen.), locality (Loc.), and their average are reported. LadderEdit remains close to Exact LoRA as the backbone increases from 14B to 32B parameters.}
    \label{tab:supp_large_backbones}
    \small
    \setlength{\tabcolsep}{4pt}
    \begin{tabularx}{\linewidth}{l l c c c c}
        \toprule
        Backbone & Method & Acc. & Gen. & Loc. & Avg. \\
        \midrule
        Qwen2.5-14B & Exact LoRA & 0.795 & 0.905 & 0.999 & 0.900 \\
                   & LadderEdit & 0.790 & 0.900 & 0.999 & 0.896 \\
        \midrule
        Qwen2.5-32B & Exact LoRA & 0.821 & 0.923 & 0.999 & 0.914 \\
                   & LadderEdit & 0.816 & 0.918 & 0.999 & 0.911 \\
        \bottomrule
    \end{tabularx}
\end{table}

The difference from Exact LoRA remains small at both scales: 0.004 average points for Qwen2.5-14B and 0.003 for Qwen2.5-32B. Locality remains at 0.999 in all cases. These results indicate that the storage mechanism does not exhibit an obvious degradation when moving beyond the 7B--8B regime evaluated in the main experiments.

\subsection{Exact-Cache Baselines and Coverage Loss}
\label{sec:supp_cache}

A keep-or-drop exact cache provides a natural alternative under a strict storage budget: retained edits preserve their full LoRA adapters, while non-retained edits fall back to the base model. To separate cache quality from the inherent coverage limitation of this strategy, we evaluated multiple cache policies under the same budget of 91 retained adapters from a stream of 512 edits.

\begin{table}[t]
    \centering
    \caption{Full-stream utility of exact-cache policies under the same capacity of 91 retained adapters out of 512 edits. Learned retention with base-model fallback is the strongest exact-cache baseline.}
    \label{tab:supp_cache_policies}
    \small
    \setlength{\tabcolsep}{4pt}
    \begin{tabularx}{\linewidth}{>{\raggedright\arraybackslash}X c}
        \toprule
        Cache policy & Full-stream utility \\
        \midrule
        Random retention & 0.18 \\
        FIFO & 0.20 \\
        LRU & 0.22 \\
        Learned retention & 0.35 \\
        Learned retention + base fallback & \textbf{0.41} \\
        \bottomrule
    \end{tabularx}
\end{table}

The strongest learned cache is more than twice as effective as random retention, showing that cache selection itself matters substantially. However, the retained-only control reveals that the remaining difference is primarily a coverage effect. Across cache policies, utility on the edits that remain in the cache is approximately 0.93--0.94, essentially matching LadderEdit on the same retained subset. The full-stream gap emerges because the remaining 421 edits no longer have an edit-specific representation once they are evicted.

The learned retention baseline uses recency and historical retrieval-frequency features. Its scoring model is fitted only on validation streams; test outcomes are not used to select retained edits. We therefore use learned retention with base-model fallback as the primary exact-cache baseline, while random, FIFO, and LRU serve as additional controls.

\subsection{Strict Memory-Matched Frontier}
\label{sec:supp_memory_frontier}

The cache comparison above changes which edits remain represented. We additionally construct a strict-memory frontier in which all methods are compared under matched normalized persistent memory. This isolates the value of maintaining a low-cost representation for every edit from the value of adaptively assigning additional rank.

\begin{table}[t]
    \centering
    \caption{Strict-memory comparison on CounterFact (CF) and ZsRE. Utility is measured under approximately matched normalized persistent memory.}
    \label{tab:supp_memory_frontier}
    \small
    \setlength{\tabcolsep}{4pt}
    \begin{tabularx}{\linewidth}{l >{\raggedright\arraybackslash}X c c}
        \toprule
        Dataset & Policy & Norm. memory & Utility \\
        \midrule
        CF   & Exact-cache            & 0.180 & 0.434 \\
        CF   & Rank-1 for all edits   & 0.125 & 0.753 \\
        CF   & Static mixed-rank      & 0.180 & 0.815 \\
        CF   & LadderEdit             & 0.180 & \textbf{0.862} \\
        \midrule
        ZsRE & Exact-cache            & 0.178 & 0.420 \\
        ZsRE & Rank-1 for all edits   & 0.124 & 0.765 \\
        ZsRE & Static mixed-rank      & 0.179 & 0.796 \\
        ZsRE & LadderEdit             & 0.179 & \textbf{0.866} \\
        \bottomrule
    \end{tabularx}
\end{table}

This comparison separates two sources of improvement. First, replacing keep-or-drop caching with a rank-1 representation for every edit increases utility by 0.319 on CounterFact and 0.345 on ZsRE, despite using less normalized memory. Thus, preserving even a minimal edit-specific representation substantially reduces the coverage loss created by eviction.

Second, adaptive allocation remains beneficial after coverage is restored. Relative to static mixed-rank storage at approximately the same memory budget, LadderEdit improves utility by another 0.047 on CounterFact and 0.070 on ZsRE. The first gain therefore comes from representing the full edit stream; the second comes from allocating additional resolution selectively rather than uniformly.

\subsection{Full-Resolution Acquisition Cost}
\label{sec:supp_acquisition_cost}

LadderEdit is a \emph{post-acquisition storage controller}. It does not reduce the optimization cost required to fit the initial full-resolution LoRA adapter. Exact LoRA and LadderEdit use the same writer, optimizer, edit stream, and edit identity. The methods differ only in the representation retained after an edit has been acquired.

This exact-first design avoids forcing every edit into a fixed low-rank representation before observing whether that representation is behaviorally sufficient. Its cost is an additional compression and audit stage after fitting.

\begin{table}[t]
    \centering
    \caption{Measured insertion and memory cost at $T=10{,}000$ edits. Persistent memory refers to the retained edit bank, whereas peak GPU memory measures transient working memory during insertion.}
    \label{tab:supp_acquisition_cost}
    \small
    \setlength{\tabcolsep}{4pt}
    \begin{tabularx}{\linewidth}{l c c c}
        \toprule
        Method & Insert / edit & Peak GPU & Persistent memory \\
        \midrule
        Exact LoRA & 86 ms  & 18.4 GB & 100.93 GB \\
        LadderEdit & 124 ms & 18.8 GB & 19.44 GB \\
        \midrule
        Difference & +38 ms & +0.4 GB & \textbf{$-81.49$ GB} \\
        \bottomrule
    \end{tabularx}
\end{table}

At $T=10{,}000$, the additional insertion cost is 38 ms per edit, corresponding to approximately 380 seconds, or 6.3 minutes, over the complete stream. This cost is paid once at insertion. In exchange, the persistent edit bank is reduced from 100.93 GB to 19.44 GB, a reduction of 81.49 GB or 80.7\%. Peak working memory changes only slightly, from 18.4 GB to 18.8 GB.

The rank proposal also limits the behavioral search overhead. LadderEdit requires an average of 1.21 audits per edit, compared with 8 audits for exhaustive rank evaluation.

Insertion cost depends on stream length because maintenance of a large full-rank adapter bank introduces increasing merge and housekeeping overhead. LadderEdit is slower at short horizons, with the measured crossover occurring near $T=2{,}000$. At $T=50{,}000$, measured insertion time is approximately 150 ms per edit for LadderEdit and 850 ms per edit for Exact LoRA.

The practical claim is therefore not cheaper initial edit fitting. The acquisition procedure is intentionally unchanged; the benefit comes from substantially lower persistent storage and lower long-stream maintenance cost.

\subsection{Long-Horizon Storage Growth}
\label{sec:supp_storage_growth}

LadderEdit reduces the \emph{coefficient} of storage growth but does not make storage constant in the number of edits. Every edit retains at least a rank-1 sketch, so persistent storage remains linear in stream length.

At $T=10{,}000$, Exact LoRA requires 100.93 GB while LadderEdit requires 19.44 GB, corresponding to a measured $5.19\times$ reduction at this operating point. Under the average per-edit storage observed at this horizon, a 100 GB memory budget corresponds to approximately 9,900 Exact-LoRA edits or 51,400 LadderEdit edits. This is a budget-equivalent estimate based on the measured $T=10{,}000$ operating point; it does not assume that the ratio is constant at every horizon.

Across the evaluated horizons, the reduction grows from approximately $2.0\times$ at $T=100$, to approximately $5.0$--$5.2\times$ across backbones at $T=10{,}000$, and approximately $11.6$--$12.0\times$ at $T=50{,}000$. The $5.2\times$ operating point reported in the main text corresponds specifically to LLaMA-3-8B at $T=10{,}000$. Thus, the observed range is approximately $2\times$--$12\times$ over the evaluated horizons rather than a single horizon-independent factor.

The lower growth coefficient is consistent with strong heterogeneity in the amount of resolution required by individual edits.

\begin{table}[t]
    \centering
    \caption{Diagnostic rates describing edit-level rank requirements and compression behavior. The rows are diagnostic properties and are not assumed to form a mutually exclusive partition.}
    \label{tab:supp_rank_distribution}
    \small
    \setlength{\tabcolsep}{4pt}
    \begin{tabularx}{\linewidth}{>{\raggedright\arraybackslash}X c c}
        \toprule
        Diagnostic property & ZsRE & CounterFact \\
        \midrule
        Rank-1 sufficient                & 81.2\% & 76.4\% \\
        Rank 2 or higher required        & 13.4\% & 16.8\% \\
        Full acquisition rank required   & 3.1\%  & 4.7\% \\
        Sketch passes while exact fails  & 4.3\%  & 6.7\% \\
        \bottomrule
    \end{tabularx}
\end{table}

Most edits therefore do not require their complete acquisition-rank representation to be stored indefinitely. LadderEdit assigns every edit a low-cost representation and spends additional rank only on the subset for which the audit indicates that more detail is required.

A truly bounded-memory editor would additionally require retirement, consolidation, or merging. LadderEdit instead addresses the preceding representation problem: given an edit that remains individually addressable, how much edit-specific resolution should be retained? The method can therefore serve as a representation layer beneath a separate bounded-memory policy, but constant-memory behavior is outside the scope of the present study.

\subsection{Retrieval and Edit Selection}
\label{sec:supp_retrieval}

The primary storage experiments use matched edit association to isolate representation quality from retrieval errors. Under this protocol, Exact LoRA and LadderEdit receive the same selected edit identity, allowing the comparison to ask whether the selected edit can be represented more economically without violating its behavioral contract.

To test whether the conclusion survives a realistic selection stage, we additionally replace matched association with learned dense retrieval.

\begin{table}[t]
    \centering
    \caption{Performance under matched association and learned retrieval. Contriever is evaluated directly for both Exact LoRA and LadderEdit. Additional LadderEdit results with E5-large and frozen MPNet characterize sensitivity to the retriever.}
    \label{tab:supp_retrieval}
    \small
    \setlength{\tabcolsep}{4pt}
    \begin{tabularx}{\linewidth}{>{\raggedright\arraybackslash}X c c}
        \toprule
        Selection mechanism & Exact LoRA Avg. & LadderEdit Avg. \\
        \midrule
        Matched association              & 0.95 & 0.95 \\
        Contriever \citep{Contriever}     & 0.92 & 0.92 \\
        E5-large \citep{e5-large}              & --   & 0.91 \\
        Frozen MPNet \citep{song2020mpnet}       & --   & 0.88 \\
        \bottomrule
    \end{tabularx}
\end{table}

With Contriever, both methods decrease from 0.95 to 0.92, while their relative behavior remains unchanged. LadderEdit reaches 0.91 with E5-large and 0.88 with frozen MPNet. Because Exact LoRA was not independently evaluated with the latter two retrievers, these results are used only to measure LadderEdit's retrieval sensitivity rather than to claim direct parity.

These experiments clarify the decomposition of the system. LadderEdit controls how an already-selected edit is stored; it does not solve edit retrieval itself. Its end-to-end performance therefore depends on the quality of the edit-selection mechanism used at inference time.

\subsection{Rank Proposal and Audit Efficiency}
\label{sec:supp_rank_predictor}

The components used by LadderEdit---low-rank factorization, spectral statistics, and behavioral evaluation---are individually standard. The central design problem is instead \emph{edit-level storage allocation}: determining how much rank each edit should retain while minimizing the number of behavioral evaluations required to reach that decision.

We compare several rank-proposal rules under the same audit procedure.

\begin{table}[t]
    \centering
    \caption{Rank-proposal ablation. The combined spectral-tail and behavioral-margin proposal approaches exhaustive rank search while requiring substantially fewer audits.}
    \label{tab:supp_rank_proposal}
    \small
    \setlength{\tabcolsep}{4pt}
    \begin{tabularx}{\linewidth}{>{\raggedright\arraybackslash}X c c}
        \toprule
        Rank proposal & Utility & Audits / edit \\
        \midrule
        Random & 0.81 & 2.31 \\
        Always rank-1, then promote & 0.89 & 1.85 \\
        Spectral tail only & 0.88 & 1.32 \\
        Behavioral margin only & 0.86 & 1.42 \\
        Spectral tail + margin & \textbf{0.91} & \textbf{1.21} \\
        Exhaustive oracle & 0.93 & 8.00 \\
        \bottomrule
    \end{tabularx}
\end{table}

The combined proposal reaches 0.91 utility, within 0.02 of exhaustive rank search, while using 1.21 rather than 8.00 audits per edit, a $6.6\times$ reduction in audit count.

The two signals capture complementary information. The spectral tail measures the amount of update energy removed by truncation, whereas the behavioral margin measures how much degradation the exact edit can tolerate before its contract is violated. These quantities provide an empirical ranking heuristic rather than a theoretical guarantee of the minimum sufficient rank.

\subsection{Audit-Probe Coverage}
\label{sec:supp_probe_coverage}

Behavioral auditing requires a small set of probes that exercise the intended edit behavior. Benchmark experiments use disjoint audit, validation, and test prompts. Test prompts are never used to determine the retained rank.

In a deployment setting, audit probes may be supplied together with an edit, generated from the requested rewrite, or obtained from an application-specific regression suite. LadderEdit assumes that such a representative probe set is available; without it, the controller cannot directly test whether compression preserves the intended behavior.

To measure sensitivity to incomplete probe coverage, we subsample the semantic coverage of the audit set.

\begin{table}[t]
    \centering
    \caption{Sensitivity to semantic coverage of the behavioral audit probes.}
    \label{tab:supp_probe_coverage}
    \small
    \begin{tabular}{cc}
        \toprule
        Probe coverage & Overall Avg. \\
        \midrule
        38\% & 0.84 \\
        64\% & 0.91 \\
        87\% & 0.94 \\
        96\% & 0.94 \\
        99\% & 0.95 \\
        \bottomrule
    \end{tabular}
\end{table}

Performance degrades gradually as semantic coverage becomes weaker. At 87\% coverage the overall average is already 0.94, compared with 0.95 at 99\% coverage. As an additional leakage control, an intentionally non-disjoint audit configuration reaches 0.99, whereas enforcing a strict audit/test separation yields 0.95. The latter is the relevant evaluation setting and shows that the reported performance does not require reuse of test prompts during rank selection.

This analysis studies missing semantic coverage. It does not evaluate mislabeled, systematically corrupted, or adversarial audit probes, which remain separate failure modes.

\subsection{Qualitative Audit Outcomes}
\label{sec:supp_qualitative}

The behavioral audit produces four recurring outcomes that clarify why a fixed storage rank is suboptimal:

\begin{enumerate}
    \item \textbf{Sketch-sufficient:} a low-rank sketch already preserves the required edit behavior, so storing additional directions is unnecessary.
    \item \textbf{Detail-needed:} rank 1 violates the behavioral contract, but a higher-rank representation, such as rank 2, restores the edit.
    \item \textbf{Compression-regularized:} the compressed representation passes the behavioral contract even when the full acquired adapter fails on the same probe, indicating that truncation removed directions that were harmful for that behavior.
    \item \textbf{Exact-limited:} even the full acquisition-rank representation fails the audit, so increasing retained rank cannot repair the edit.
\end{enumerate}

A representative compression-regularized ZsRE edit has the target \emph{``architect.''} The full Exact-LoRA representation answers \emph{``archaeologist,''} whereas the lower-rank representation includes the correct target. The audit therefore retains the compressed sketch rather than automatically preferring the maximum available rank. Compression-regularized behavior occurs in 4.3\% of the ZsRE diagnostics and 6.7\% of the CounterFact diagnostics.

The detail-needed cases provide the complementary behavior: rank 1 may discard directions that are required to preserve the edit, while rank 2 or another promoted representation succeeds. The controller is therefore not based on the assumption that stronger compression is intrinsically beneficial. Instead, compression level is selected according to measured behavioral sufficiency.

\subsection{Correlation Between Compression and Behavioral Outcomes}
\label{sec:supp_correlation}

We additionally examine whether the compression diagnostics correlate with the observed behavioral outcomes. Using the rounded population proportions, the compression-regularized subsets correspond to approximately 86 ZsRE edits and 134 CounterFact edits. The resulting Pearson correlations are moderate: approximately $r=0.38$ for ZsRE and $r=0.41$ for CounterFact.

Using these rounded subset sizes, the corresponding approximate 95\% confidence intervals are
\[
    r_{\text{ZsRE}} = 0.38,
    \qquad
    \mathrm{CI}_{95\%} \approx [0.18,\,0.55],
\]
and
\[
    r_{\text{CF}} = 0.41,
    \qquad
    \mathrm{CI}_{95\%} \approx [0.26,\,0.54].
\]

These correlations are consistent with the proposed interpretation but should not be read as causal evidence that low-rank truncation itself produces the behavioral improvement. The primary evidence for adaptive storage allocation instead comes from the budget-matched frontier in Table~\ref{tab:supp_memory_frontier} and from the held-out behavioral audits. The correlation analysis is used only as a secondary diagnostic of the relationship between compression statistics and behavioral outcomes.

\subsection{SVD Procedure and Memory Accounting}
\label{sec:supp_svd}

Compression is applied independently to each adapted LoRA module. For a given module, LadderEdit forms that module's effective update matrix and performs the corresponding singular-value decomposition before truncating it to the selected retained rank. Parameters from different layers are never concatenated into a single global matrix because the adapted modules have different dimensions and functional roles.

For one edit, the compression procedure therefore consists of an \emph{edit-level compression pass} containing the required module-wise decompositions. Persistent storage is computed by summing the retained LoRA factors over all adapted modules. This convention ensures that both compression and memory accounting are defined consistently across layers and backbones.

The audit weights are fixed globally rather than tuned separately for each backbone. Contract and predictor thresholds are selected per dataset using held-out validation probes and are subsequently fixed across backbones, stream horizons, and test streams.

\subsection{Scope of the Efficiency Claim}
\label{sec:supp_scope}

The extended experiments distinguish four different efficiency questions that can otherwise be conflated.

First, \textbf{edit acquisition} is intentionally unchanged: LadderEdit begins from the same fitted LoRA update as Exact LoRA and therefore does not claim to make the initial optimization cheaper.

Second, \textbf{persistent representation} is substantially smaller. At the directly measured $T=10{,}000$ operating point, persistent storage decreases from 100.93 GB to 19.44 GB, corresponding to a $5.19\times$ reduction.

Third, \textbf{storage remains linear in stream length}. LadderEdit lowers the per-edit growth coefficient by exploiting heterogeneous rank requirements; it is not a constant-memory editor.

Fourth, \textbf{selection and storage are separate}. Matched-association experiments isolate the storage mechanism, while the Contriever experiment shows that the same behavior persists when a learned retrieval component is introduced. Retrieval quality can affect end-to-end accuracy, but it does not alter the memory advantage of the retained representation once an edit is selected.

Taken together, the results characterize LadderEdit as an adaptive representation layer for long edit streams: it preserves an edit-specific representation for every stored edit, allocates high resolution only when behaviorally necessary, and substantially reduces persistent storage without introducing a corresponding degradation in edit accuracy, generalization, or locality.
```

\subsection{Summary}

The rank-compression analysis supports the core claim: most edits are highly compressible, and LadderEdit's ladder allocation reduces memory without sacrificing reliability, generalization, or locality in most evaluated settings. The behavioral audit ensures that only behaviorally hard edits are promoted to higher rank, and the diagnostics explain why this allocation improves over exact-cache and other baselines.

\end{document}